# A Gravity-informed Spatiotemporal Transformer for Human Activity Intensity Prediction

Yi Wang, *Student Member, IEEE*, Zhenghong Wang, Fan Zhang, Chaogui Kang, Sijie Ruan, *Member, IEEE*, Di Zhu, Chengling Tang, Zhongfu Ma, Weiyu Zhang, Yu Zheng, *Fellow, IEEE*, Philip S. Yu, *Life Fellow, IEEE*, Yu Liu

***Abstract*—Human activity intensity prediction is crucial to many location-based services. Despite tremendous progress in modeling dynamics of human activity, most existing methods overlook physical constraints of spatial interaction, leading to uninterpretable spatial correlations and over-smoothing phenomenon. To address these limitations, this work proposes a physics-informed deep learning framework, namely Gravity-informed Spatiotemporal Transformer (Gravityformer) by integrating the universal law of gravitation to refine transformer attention. Specifically, it (1) estimates two spatially explicit mass parameters based on spatiotemporal embedding feature, (2) models the spatial interaction in end-to-end neural network using proposed adaptive gravity model to learn the physical constraint, and (3) utilizes the learned spatial interaction to guide and mitigate the over-smoothing phenomenon in transformer attention. Moreover, a parallel spatiotemporal graph convolution transformer is proposed for achieving a balance between coupled spatial and temporal learning. Systematic experiments on six real-world large-scale activity datasets demonstrate the quantitative and qualitative superiority of our model over state-of-the-art benchmarks. Additionally, the learned gravity attention matrix can be not only disentangled and interpreted based on geographical laws, but also improved the generalization in zero-shot cross-region inference. This work provides a novel insight into integrating physical laws with deep learning for spatiotemporal prediction.**

## I. INTRODUCTION

Human activity intensity [1-3], also known as the dynamic distribution of active population, quantifies the number of individuals locating in an area for an extended duration while engaging in identifiable activities, thereby accurately reflecting the functional significance of that place. Predicting human activity intensity across different urban places plays a critical role in optimizing urban planning [4], enhancing traffic management [5], and improving public safety [6], constituting a fundamental task in urban computing [7]. Different from conventional crowd flow prediction, human activity intensity prediction excludes transient human movements patterns and focuses on stable activity patterns, enabling more accurate and effective location-based tailored services [8, 9].

With the emergence of the big data era, deep learning-based methods have emerged as mainstream approaches in spatiotemporal prediction due to their robust nonlinear modeling capabilities [10]. These methods, exemplified by Spatiotemporal Graph Neural Networks (ST-GNNs) [11-13], been proliferated in numerous fields, such as crowd flow prediction [12], environmental pollution prediction [14], and crime prediction [15]. They are also well-suited for capturing the complex spatiotemporal dynamics inherent in human activity patterns [16, 17].

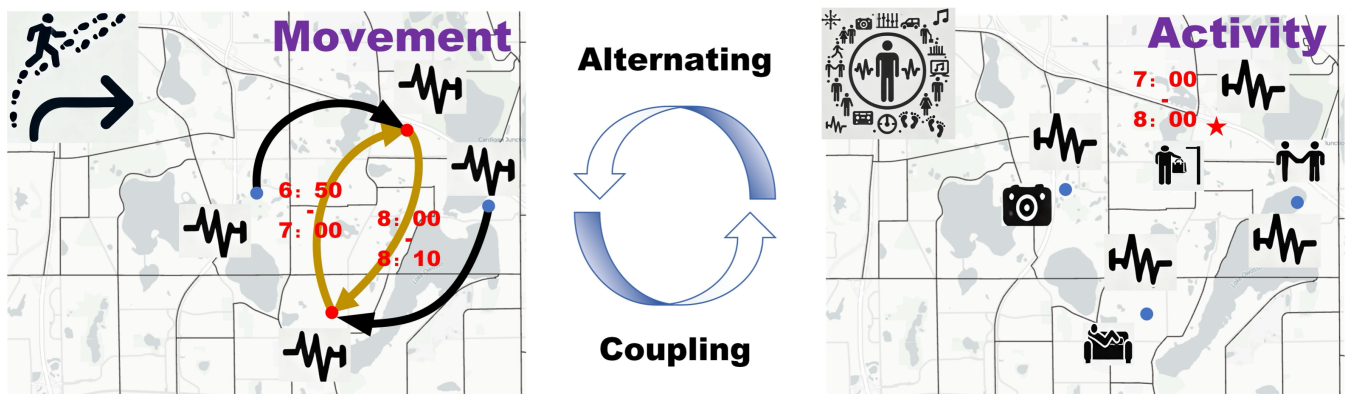

**Fig. 1** The human activity and movement are alternating and coupling. The movement from the residence to the mall occurs between 6:50 and 7:00, while the return trip from the mall to the residence takes place between 8:00 and 8:10. The shopping activity occurs from 7:00 to 8:00.

However, the prediction of human activity intensity is governed by unique spatiotemporal dynamics that distinguish it from other spatiotemporal analyses. The fundamental distinction lies in the intrinsic coupling between human movements and activities—these two phenomena alternate and mutually influence each other within urban systems. Specifically, human movement serves as the enabler of human activities: individuals must travel to specific locations to engage in intended activities, and these activities subsequently generate new movement demands. For instance (Fig. 1), individuals travel from their residence to a shopping mall (movement), engage in shopping activities (activity), then return home (movement). This movement-activity coupling

This work is supported by the National Natural Science Foundation of China (Grant # 42430106, 42371468, 424B2013). *(Yi Wang and Zhenghong Wang have equal contributions). (Corresponding author: Fan Zhang).*

Yi Wang, Zhenghong Wang, Fan Zhang, Chengling Tang, Weiyu Zhang and Yu Liu are with Institute of Remote Sensing and Geographic Information System, School of Earth and Space Sciences, Peking University, Beijing 100871, China. (e-mail: fanzhanggis@pku.edu.cn)

Chaogui Kang is with National Engineering Research Center of Geographic Information System, China University of Geosciences (Wuhan) 430074, China.

Sijie Ruan is with School of Computer Science and Technology, Beijing Institute of Technology, Beijing 100081, China.

Di Zhu and Zhongfu Ma are with Department of Geography, Environment and Society, University of Minnesota, Twin Cities, Minneapolis, MN 55455, USA.

Yu Zheng is with JD iCity, JD Technology, Beijing 100176, China.

Philip S. Yu is with Department of Computer Science, University of Illinois Chicago, Chicago 60607, USA.

reveals a critical insight: the spatial distribution of current human movements inherently encode information about where future activities will likely occur. At the regional level, the direction and intensity of movement between different regions constitute a form of spatial interaction [18]—a special type of spatial correlation with physical significance. Therefore, accurately predicting human activity intensity requires explicitly modeling the spatial interactions that govern human movements [16], as these interactions determine potential distributions of future activity.

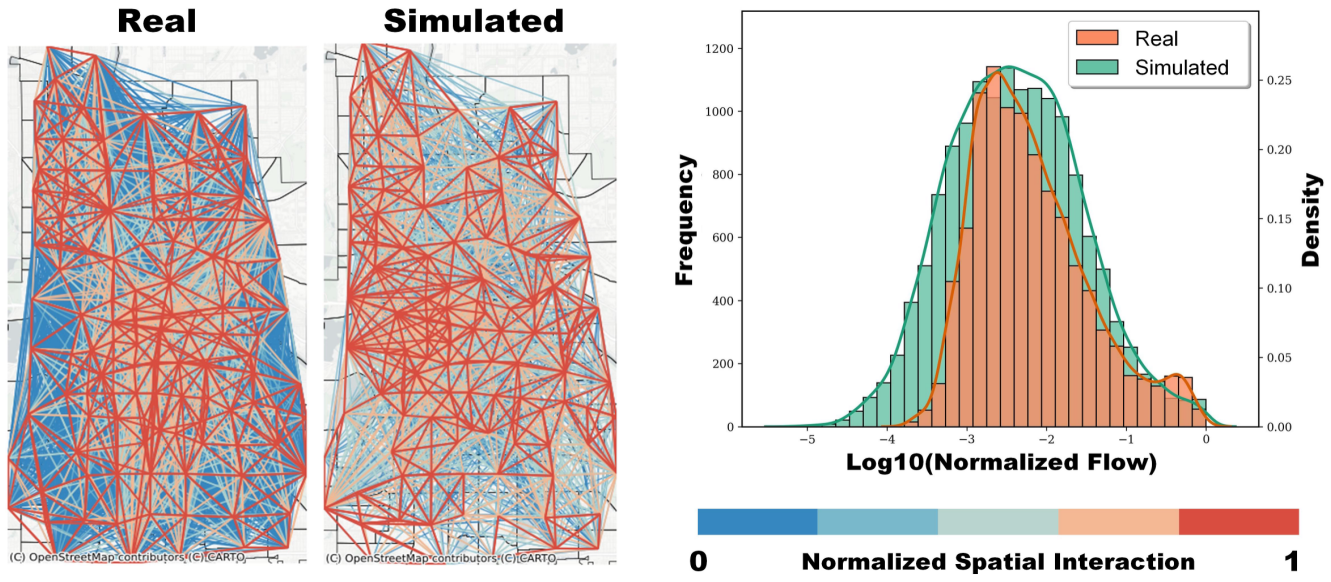

**Fig. 2** Comparison between real and simulated spatial interactions. We simulate potential spatial interactions using the gravity model [19] according to human activity intensity in each region. The simulated interactions exhibit similar spatial and numerical distributions compared to real Origin-Destination (OD) flow data.

Although existing ST-GNNs perform well in learning general spatial correlations (e.g., spatial similarity), they struggle to capture these more complex spatial interactions. Specifically, they face two critical challenges when applied to human activity intensity prediction.

- **Absence of physical constraints:** Existing data-driven methods often rely on unconstrained learning assumptions, resulting in learned spatial correlation matrices that are random and difficult to interpret [11, 20]. This is particularly problematic because the patterns of real-world human movement (i.e., spatial interaction) exhibits regularities driven by the fundamental trade-off between travel costs and destination attractiveness [21]. For example, when seeking medical care, people consider both a hospital's reputation and its distance, rather than simply defaulting to the nearest option. Fortunately, mathematical analysis has proven that these spatial interaction patterns conform to well-defined closed-form solutions [22-24]. These solutions demonstrate that the interaction intensity between locations is proportional to their attractiveness (e.g., population) and inversely proportional to the distance between them [19]. In other words, within a human movement system, the underlying spatial interaction distribution can be derived through its closed-form solution (Fig. 2). Therefore, its closed-form solution can also serve as a physical constraint for spatial correlation learning, guiding models toward realistic and interpretable spatial modeling while circumventing the need for privacy-sensitive and fine-grained individual movement data [25]. Unfortunately, the lack of understanding of these physical laws in most existing ST-GNNs complicates the capture of invariant patterns in human activity, thereby increasing prediction difficulty.
- **Over-smoothing phenomenon:** Unconstrained spatial correlation learnings are particularly susceptible to the over-smoothing problem prevalent in deep neural networks, especially transformer-based models [26-28]. Over-smoothing refers to the tendency for output features to become increasingly homogeneous as network depth increases, caused by repeated aggregation of neighborhood features [28, 29]. This results in highly sparse and redundant attention matrices in transformers (most correlation values converging toward zero), thereby reducing effective information propagation [26]. In human activity prediction, this phenomenon is especially problematic for urban-scale applications involving numerous spatial nodes [30], where over-smoothing eliminates crucial distinctions between different locations. Moreover, such homogeneous patterns fundamentally contradict the heterogeneous nature of real-world spatial interactions governed by physical laws. Consequently, the over-smoothing problem significantly reduces the efficacy of spatiotemporal predictions, particularly at large spatial scales.

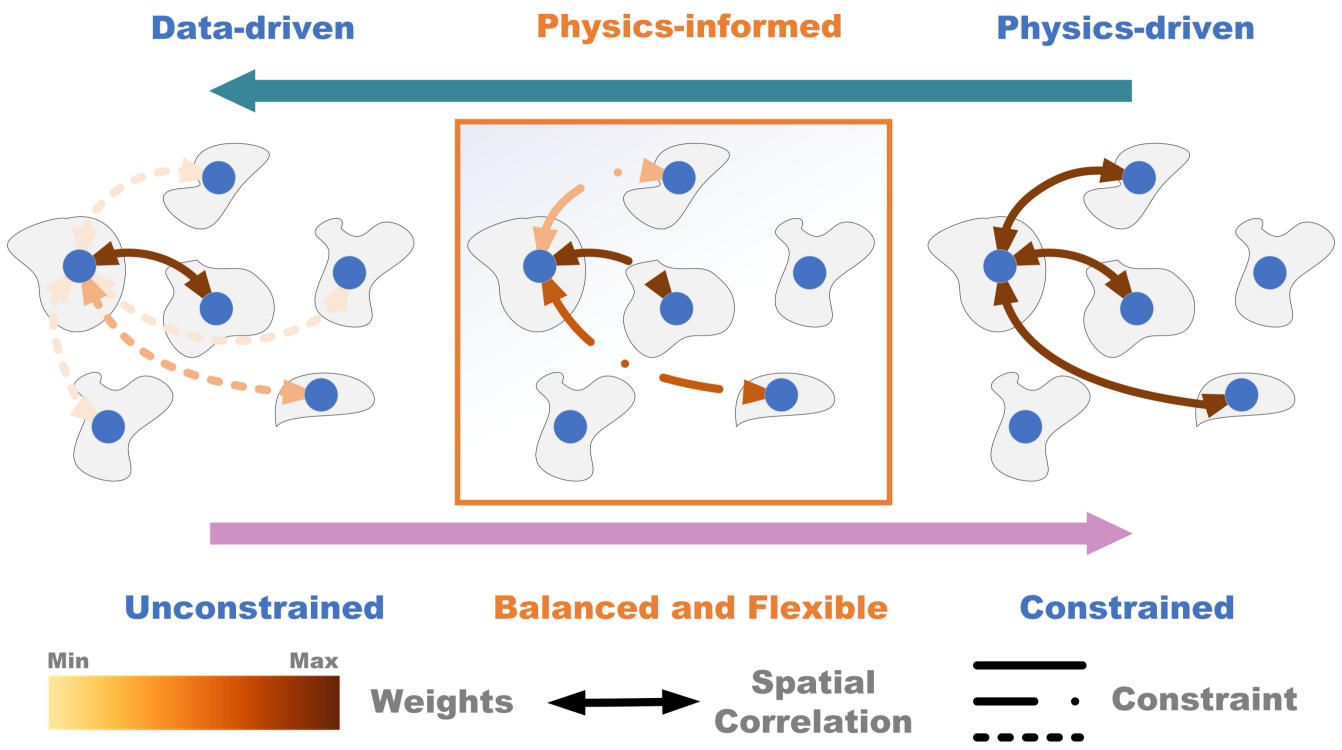

**Fig. 3** Comparison of data-driven, physics-driven and physics-informed methods on spatial correlation learning. Data-driven models rely on large datasets of observation, while physics-driven models are based on physical laws and equations. Physics-informed models use these laws to guide or constrain data-driven models, enhancing accuracy and robustness. Due to differing assumptions, physics-driven methods often have overly fixed spatial correlations, whereas data-driven methods may exhibit excessively random spatial correlations. In contrast, physics-informed methods provide a balanced and more flexible modeling.

To solve abovementioned challenges, we propose a physics-informed machine learning (PIML) model [31], namely Gravity-informed Spatiotemporal Transformer (Gravityformer). Specifically, we integrate the law of universal gravitation [19] into transformer attention for adaptive spatial interaction modeling. As illustrated in Fig. 3, our model flexibly constrains spatial correlation learning through physical law. The learned interaction matrix can not only guide transformer in deep layers to mitigate the over-smoothing phenomenon but also provides interpretable spatial relationships aligned with geographical principles. The contributions of this study are as follows:

- Based on the law of universal gravitation, we propose a gravity-informed transformer framework that spatially-explicitly capture potential interaction through physical constraints. Leveraging the closed-form solution of spatial interaction, we develop an adaptive gravity (AdaGravity) module to represent the physical constraint in end-to-end neural network. By integrating the AdaGravity with the transformer attention, we achieve an interpretable gravity attention matrix that alleviate the over-smoothing problem in deep transformer layers.
- Inspired by the natural coupling between spatial and temporal dynamics in human activity systems, we design a parallel spatiotemporal graph transformer module, namely ST-GC2former. It dynamically integrates temporal and spatial learning while maintaining balanced coupling relationships between spatiotemporal features.
- To further evaluate our Gravityformer, extensive experiments were conducted using large-scale real-world human activity data from six metropolitan areas in the United States for human activity intensity prediction. Compared with other state-of-the-art methods, our model demonstrates superior performance, outstanding interpretability, and competitive generalization across diverse urban contexts.

## II. Related Work

### *A. Spatiotemporal prediction for human activity intensity*

Previous works predominantly used statistical methods [32, 33] due to limited data, but these struggled with complex and nonlinear scenarios. With the rapid advancement of social sensing [34] and mobile communication [35] technologies, a substantial data foundation has been provided for fine-grained spatiotemporal forecasting. Capitalizing on this wealth of data, deep learning provides a new opportunity for spatiotemporal prediction. These methods, particularly ST-GNNs [11-13], integrate Graph Neural Networks (GNNs) or transformer models with time-series analysis techniques, effectively capturing both spatial and temporal features [36-40].

The evolution of these deep learning models has been significant. Early approaches often operated under the assumption of static and neighborhood-homogeneous spatial dependencies, establishing a baseline for capturing fundamental patterns [36, 41, 42]. To achieve superior performance, subsequent research focused on modeling more complex and realistic relationships, including dynamic correlations, frequency domain relationships, node-specific meta patterns, multi-scale dependencies, and spatiotemporal heterogeneity [43-48]. Then, the integration of the Transformer architecture [54] also marked another significant leap, with its self-attention mechanism inspiring a new wave of powerful variants that consistently pushed performance boundaries [38, 39, 49-51]. In addition to performance, model efficiency has emerged as a critical consideration. Recent work has addressed this by proposing lightweight architectures based on MLPs, CNNs or linearized Transformers [52-55], and techniques like time-series disentanglement to reduce parameters [56, 57]. These models have also been proliferated in other fields, such as transportation [12], environmental monitoring [14], and crime prediction [15].

Due to their robust capabilities of spatiotemporal prediction, those models have also been utilized in human activity intensity prediction. Due to the complex spatial relationships within human activities, most of studies have emphasized the significance of spatial modeling. It includes the multiple spatial interaction information [16], the high-order relationships [58] and the spatially-explicit multiplex relationships [17]. However, such approaches exhibit a limitation. They often depend on fine-grained and difficult-to-acquire origin-destination (OD) data, or they lack physics constraints governing human movement, which leads to uninterpretable learned correlation and poor generalization.

### *B. Physics-informed machine learning for spatial interaction*

As the limitations of those data-driven models become more apparent, PIML have been proposed and proliferated, such as traffic state prediction [59, 60], climate modeling [61-63] and remote sensing estimation [64, 65]. PIML combines physical knowledge with data-driven machine learning approaches to achieve superior optimization and more generalization [31]. It includes four paradigms: physics-informed loss functions, physics-informed initialization, physics-informed architecture design, residual modeling and hybrid physics machine learning models.

In spatial interaction-related applications, some classic physical models, e.g., gravity model [19] and radiation model [66] and potential field model [67], are being challenged by PIML models. Generally, those PIML models are utilized for human migration generation and trajectory prediction in spatial interaction modeling. For example, Deep Gravity [68] replaces the gravity model with a nonlinear MLP to generate more complex population migration patterns. ODGN [69] uses the gravity model as a decoder to decode features from a multi-view GAT for population migration generation. PI-NeuGODE [70] integrates physical models and symbolic regression into neural ordinary differential equations (ODEs) to discover physical knowledge from complex data and improve prediction accuracy.

In traffic flow prediction, PIML-related methods have also emerged [71-73]. Ji et al. proposed a traffic flow prediction model based on potential energy fields, namely STDEN [74]. It incorporates gravitational potential energy into neural ODEs to achieve physics-informed spatiotemporal modeling. Based on STDEN, Wang et al. decomposed complex traffic flow graphs into multiple potential fields using community division and multi-tree decomposition [20], simplifying the computational complexity. It also addresses bidirectional and cyclic flow issues. While those methods achieve high-performance, they may require finer-grained data, such as edge-based flow or OD flow, which might be unavailable due to privacy concerns, limiting their applicability.

Therefore, inspired by PIML, we propose an adaptive gravity model designed to integrate the principles of universal gravitation into an end-to-end neural network. Unlike traditional applications, our model does not generate OD flows. Instead, it serves as a prior physics knowledge, imposing a constraint on the spatial correlations learning. This approach enables to guide the spatiotemporal Transformer, mitigating the over-smoothing phenomenon and achieving an interpretable, spatially-explicit learned process.

## III. Methodology

To incorporate the physical laws of human movement into spatiotemporal prediction, Gravityformer was proposed (Fig. 4). It consists of three main components: spatiotemporal adaptive embedding, the adaptive gravity model (AdaGravity) and the gravity-informed spatiotemporal graph convolution Transformer (GST-GC2former). The following sections provide detailed descriptions of each component.

### A. Preliminary

In this study, the spatial structure of city is defined as an undirected graph $G(\mathcal{V},\mathcal{E})$, where $\mathcal{V}\in\mathbb{R}^{N}$ represents the vertices within the graph, i.e., the spatial units within the city; $\mathcal{E}\in\mathbb{R}^{N\times N}$ represents the edges within the graph, i.e., the connections among different spatial units. Then, human activity intensity prediction can be defined that: Given historical observations of human activity intensity $X_H=\{X_{t-q+1},X_{t-q+2},\dots,X_t|G\}\in\mathbb{R}^{N\times T\times C}$, crowd inflow $X_{in}=\{X_{t-q+1}^{in},X_{t-q+2}^{in},\dots,X_t^{in}|G\}\in\mathbb{R}^{N\times T\times C}$, and crowd outflow $X_{out}=\{X_{t-q+1}^{out},X_{t-q+2}^{out},\dots,X_t^{out}|G\}\in\mathbb{R}^{N\times T\times C}$ over $q$ timesteps on the predefined graph $G$, the goal is to predict the future $p$ snapshots of human activity intensity $Y_P=\{X_{t+1},X_{t+2},\dots,X_{t+p}|G\}\in\mathbb{R}^{N\times T\times C}$ using the forecasting model $\mathcal{F}$.

$$\{X_H,X_{in},X_{out}|G\}\underset{\mathcal{F}}{\rightarrow}Y_P=\{X_{t+1},X_{t+2},\dots,X_{t+p}|G\},\tag{1}$$

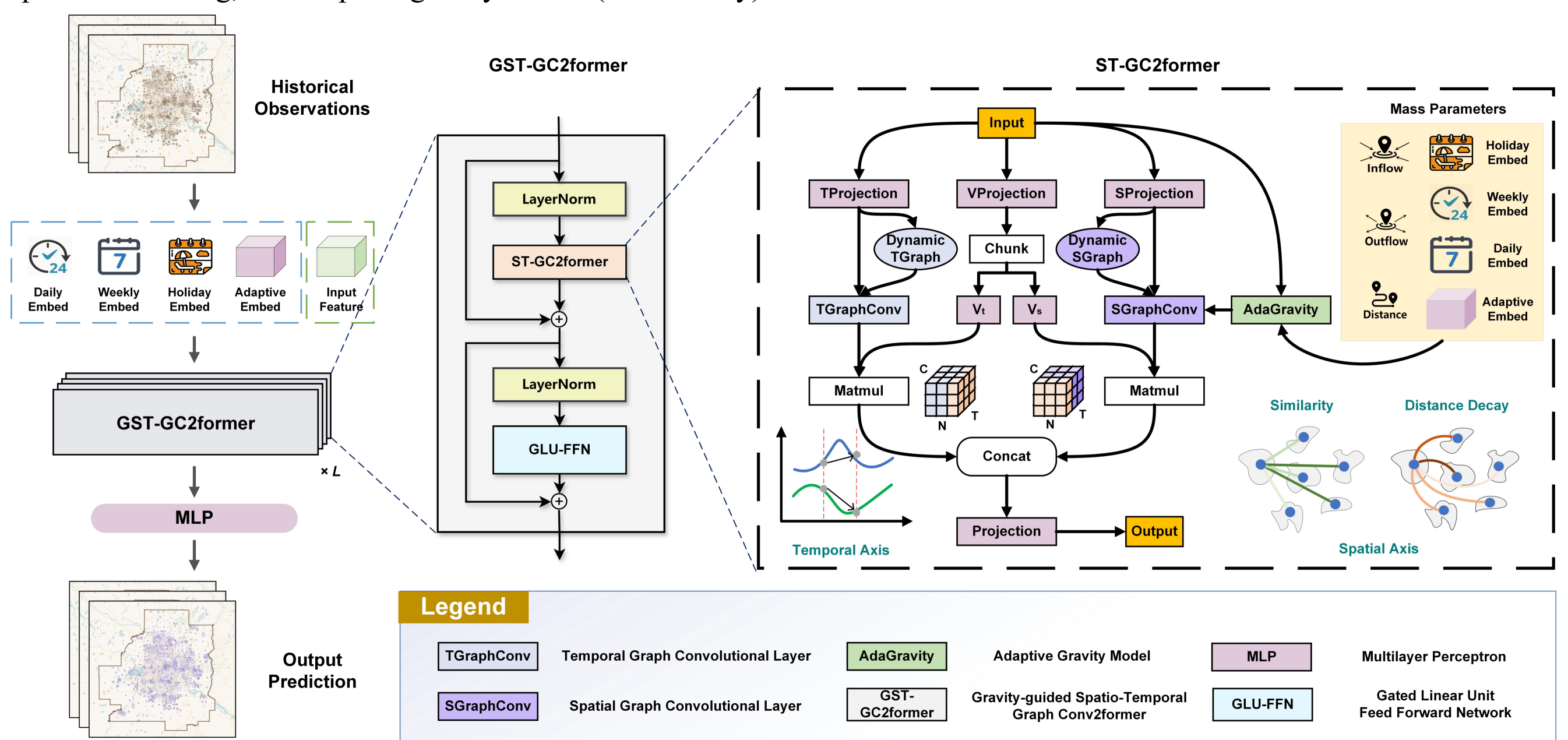

**Fig. 4** Overall framework of proposed Gravityformer.

### B. Spatiotemporal adaptive embedding

In time-series analysis, temporal periodicity and holiday effect are two of the most crucial characteristics [75] that describe the temporal patterns within the data. Effectively representing these patterns is essential for understanding spatiotemporal data.

For temporal periodicity, two learnable embeddings are defined: the time-of-day embedding $E_d\in\mathbb{R}^{D\times C_t}$ and the day-of-week embedding $E_w\in\mathbb{R}^{7\times C_t}$. To address abrupt changes caused by holiday effects (e.g., Christmas and Thanksgiving), a learnable holiday embedding is also defined, i.e., the holiday embedding $E_h\in\mathbb{R}^{2\times C_t}$. These embeddings represent the meta timestamp information within sequence, such as time-of-day $S^d\in\mathbb{R}^{T}$, day-of-week $S^w\in\mathbb{R}^{T}$, and holiday status $S^h\in\mathbb{R}^{T}$, including the details like 14:00, Monday and non-holiday. Then, an adaptive spatiotemporal embedding $E_a\in\mathbb{R}^{N\times T\times C_a}$ is utilized to capture the complex potential spatiotemporal relationships within the data [39]. This sequence-shared and assumption-free learnable parameter adjusts adaptively to enhance the learning of inherent spatiotemporal patterns.

Additionally, to represent the raw information $X\in\mathbb{R}^{N\times T\times C}$ from historical data, a simple fully connection (FC) layer is employed to obtain the spatiotemporal features $E_f \in \mathbb{R}^{N \times T \times C_f}$. By concatenating all the embedding features, the final spatiotemporal representation was available $Z \in \mathbb{R}^{N \times T \times C_{in}}$, which serves as the input to the GST-GC2former Block:

$$E_f = W_f X + b_f, \tag{2}$$

$$Z = W_o(E_d || E_w || E_h || E_a || E_f) + b_o, \tag{3}$$

where $W_f \in \mathbb{R}^{C \times C_f}$, $b_f \in \mathbb{R}^{C_f}$, $W_o \in \mathbb{R}^{(3C_t + C_a + C_f) \times C_{in}}$, $b_o \in \mathbb{R}^{C_{in}}$ are learnable parameters in FC for changing the feature dimension.

### *C. Adaptive gravity model*

The spatial interaction information generated by human movement contains essential physical laws often overlooked in node-oriented spatiotemporal prediction. To explicitly consider the constraint of closed-form solution in human spatial interaction systems, the law of universal gravitation [76] was incorporated into our model. Proposed by Newton in physics [77], this law states that any two objects experience an attractive force $F$, directly proportional to the masses $m_i$ and $m_j$ of the objects and inversely proportional to the distance $r$ between them (Eq. 4).

$$F = G \frac{m_i m_j}{r^2}, \tag{4}$$

where $G$ is the gravity constant. With the development in human movement research, the gravity model has been proposed motivated based on such law [19], facilitating the estimation of population migration between two locations. To enhance the fitting capability of the gravity model, some learnable parameters $G, \alpha_1, \alpha_2$ have been introduced to scale the mass and distance parameters. It can be expressed as follows:

$$T_{ij} = G \frac{{P_i}^{\alpha_1} {P_j}^{\alpha_2}}{d_{ij}^{\beta}}, \tag{5}$$

where $T_{ij} \in \mathbb{R}^{N \times N}$ represents the connection strength of spatial interaction between two locations, and $d_{ij} \in \mathbb{R}^{N \times N}$ represents the distance between them. The input variables $P_i$ and $P_j \in \mathbb{R}^{N \times 1}$ denote the populations (or other relevant data) of the two locations. The parameters $\alpha_1$ and $\alpha_2$ are scaling factors for mass, while $\beta$ is a scaling factor for distance.

Unfortunately, gravity models have traditionally been applied to semantically interpretable and low-dimensional attributes. In its traditional applications in economics and geography, the model computes spatial interactions using single meaningful attribute [78] (e.g., gross domestic product). When handling multiple features, they often employ a decomposed approach [79], calculating interactions for each modality and linearly summing the results. This methodology is incompatible with the high-dimensional tensors from neural networks, whose latent dimensions are semantically opaque. As attributes are encoded across the high-dimensional space, a direct dot product for the mass vector weakens the interaction magnitude while simultaneously introducing semantically meaningless and redundant similarities.

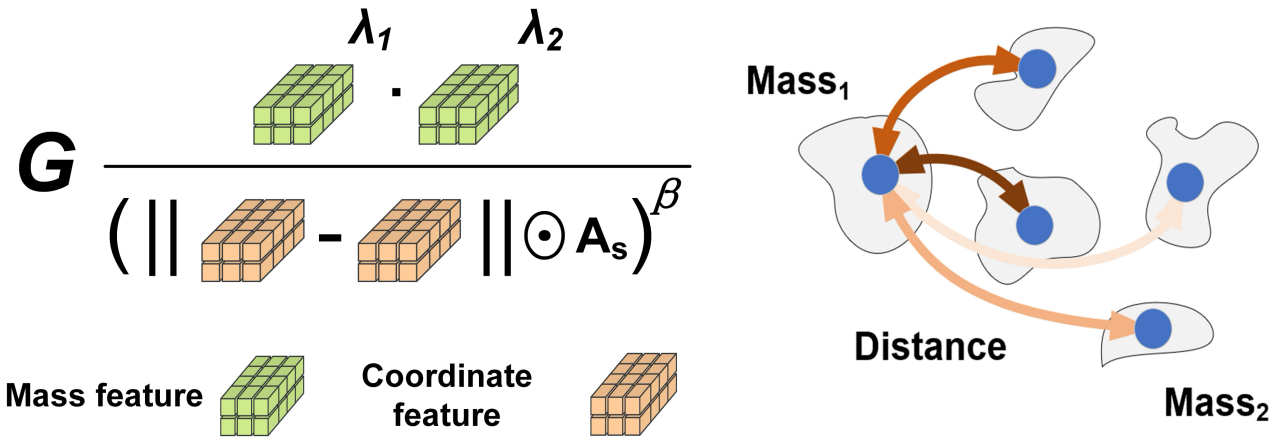

**Fig. 5** The architecture of the proposed AdaGravity module. The left subfigure illustrates the formulation of the proposed adaptive gravity model, while the right subfigure depicts a schematic representation of the generated spatial interaction.

To solve such challenges, a simple yet effective module is proposed (Fig. 5), namely adaptive gravity model (AdaGravity), for learning the potential spatial interaction among nodes from physics perspective. The central idea of our approach is to derive a scalar mass parameter for each node by aggregating its high-dimensional and learnable feature tensor. This design choice is motivated by two key factors. First, since the gravity model evaluates the magnitude of interaction rather than feature similarity, a scalar mass aligns more closely with the physical intuition of gravity model and enhances its interpretability. Second, optimizing a magnitude representation of scalar parameter is a significantly more tractable and stable problem than optimizing a magnitude representation of multi-dimensional vector. This is empirically validated in our ablation study (Section V.A).

Based on this principle, we chose average pooling as the aggregation function. Its simplicity and computational efficiency make it a suitable choice, while it also mitigates meaningless similarities in high-dimensional space. This approach is also analogous to the decomposed gravity models [79] used in traditional econometrics, where interactions are aggregated across different sectors. Specifically, the mass parameters $M_i$ and $M_j$ are computed dynamically within each layer $L$. This calculation integrates the updated embedding features $Z^{(l)}$ and initial embedding features $Z$, with external inflow and outflow data. The inclusion of inflow and outflow is crucial for creating a better representation of spatial interaction, specifically by better differentiating each node's capacity for attraction (inflow) and repulsion (outflow). To ensure dimensional compatibility for this integration, the inflow and outflow features are first projected into the same latent space as $Z^{(l)}$ via two independent FC layers, resulting in the embeddings $E_{in}$ and $E_{out}$. Then, the representation of the mass parameters is shown as Eq.6:

$$E_{in} = W_{in} X_{in} + b_{in}, E_{out} = W_{out} X_{out} + b_{out}, \tag{6}$$

$$M_i = Avg\left(E_{in} || Z || Z^{(l)}\right), M_j = Avg\left(E_{out} || Z || Z^{(l)}\right), \tag{7}$$

where $E_{in}$ and $E_{out} \in \mathbb{R}^{N \times T \times C_f}$ represent the hidden features for inflow and outflow, while $W_{in}$ and $W_{out} \in \mathbb{R}^{1 \times C_f}$ and $b_{in}$ and $b_{out} \in \mathbb{R}^{C_f}$ are learnable parameters of FC layers. The $Avg(\cdot)$ denotes average pooling for obtaining explicit mass parameters. It adheres to the scalar form of the classical gravity model, rendering mass parameters explicit and analyzable. Furthermore, the integration of the initial features $Z$ stabilizes the mass parameter updates, reducing over-smoothing caused by overly large or small updated feature.

For the distance $d_{ij}$, ST-GNNs typically employ gaussian kernel functions to account for the distance decay effect [36]. Similarly, we also use this approach by incorporating the processed distance matrix $A_d \in \mathbb{R}^{N\times N}$ into AdaGravity, thus mitigating gradient explosion issues caused by zero distances. However, geographical distance represents only a simplistic depiction of node association [80], which may not be optimal for prediction tasks. To address the complex potential distances in the feature space, adaptive distance scaling is introduced into our model. Inspired by flow-based cartogram techniques [81], the distance matrix can be further scaled while maintaining topological invariance, thereby increasing adaptive capability. Specifically, an adaptive undirected graph is utilized to weight the distance matrix for identifying the most suitable potential distance between nodes.

$$E_{s1} = \rho(\kappa S_1), E_{s2} = \rho(\kappa S_2), \tag{8}$$

$$A_s = \phi(\rho\left(\kappa\left(E_{s1}{E_{s2}}^{\intercal} - {E_{s1}}^{\intercal}E_{s2}\right)\right)), \tag{9}$$

where $E_{s1}$ and $E_{s2} \in \mathbb{R}^{N\times C_e}$ denotes the adaptive node vector, $\kappa$ denotes the hyperparameter for controlling the saturation rate, $\rho(\cdot)$ and $\phi(\cdot)$ denotes the Tanh activation function and ReLU activation function, respectively. Then, the scaling matrix $A_s$ includes all scaling factor $s_{ij}$.

Finally, our AdaGravity can be expressed as follow:

$$\begin{aligned} A_{ag} &= \sigma(G) \odot \frac{{M_i}^{\sigma(\alpha_1)}{M_j}^{\sigma(\alpha_2)}}{\left(d_{ij} \odot \frac{1}{s_{ij}}\right)^{\sigma(\beta)}}, \\ &= \sigma(G) \odot \left({M_i}^{\sigma(\alpha_1)}{M_j}^{\sigma(\alpha_2)}\right) \odot (A_d \odot A_s)^{\sigma(\beta)}, \end{aligned} \tag{10}$$

where $\sigma(\cdot)$ denotes the Softplus activation function for ensuring the non-negativity of all learnable parameters $G, \alpha_1$ and $\alpha_2$ thereby preserving the form of the gravity model. Due to its dynamic updating and shallow representation, it can provide physics constraint to the proposed model for alleviating the over-smoothing phenomenon.

*D. Gravity-informed Spatiotemporal Graph Convolution Transformer*

*(1) Parallel Spatiotemporal Transformer Module*

Existing transformer-based frameworks, like ASTGNN [38] and STAEformer [39], traditionally model temporal and spatial patterns separately. However, in dynamically changing human activity data, temporal and spatial patterns are dynamically coupled. For instance, human activity destination can be influenced by multiple locations over multiple time steps. To address this issue, a parallel spatiotemporal graph convolution transformer was proposed, namely ST-GC2former, for achieving a balanced integration of spatial and temporal learning.

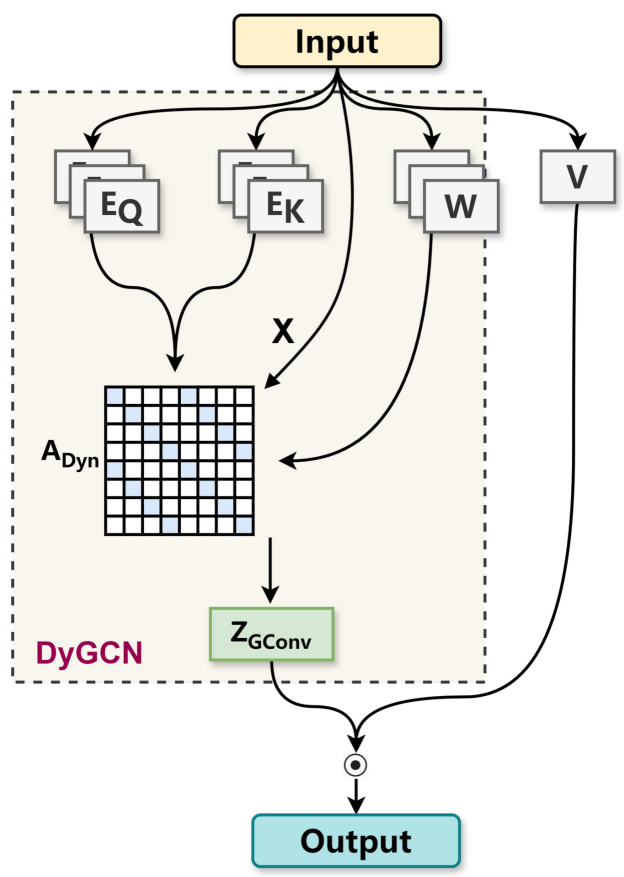

**Fig. 6** The architecture of the proposed GC2former module.

Specifically, the proposed model comprises two parallel transformer branches, with one branch illustrated in Fig. 6. Given the structural similarity between transformer and graph convolution network, transformer can be also viewed as a dynamic graph convolution layer [28, 38]. Inspired by Conv2former [82], a new branch between $V_T$ and $V_S \in \mathbb{R}^{N\times T\times C_d}$ is introduced into the dynamic graph convolution (DyGCN) for dynamic feature adjustment. This branch facilitates synchronized feature adjustments across temporal and spatial dimensions in the DyGCN layers, enabling differential coupling learning from diverse geographical locations over time. Since DyGCN is considered as the convolution layer of Conv2former, we employed a single-head attention projection. Specifically, the proposed ST-GC2former can be represented as follows:

$$E_{Q_t} = Z^{(l-1)}W_{Q_t}, E_{K_t} = Z^{(l-1)}W_{K_t}, \tag{11}$$

$$E_{Q_s} = Z^{(l-1)}W_{Q_s}, E_{K_s} = Z^{(l-1)}W_{K_s}, \tag{12}$$

$$V_T = Z^{(l-1)}W_{V_t}, V_S = Z^{(l-1)}W_{V_s}, \tag{13}$$

$$A_T = \frac{E_{Q_t}{E_{K_t}}^{\intercal}}{\sqrt{C_d}}, A_S = \frac{E_{Q_s}{E_{K_s}}^{\intercal}}{\sqrt{C_d}} \tag{14}$$

$$Z_T' = \emptyset(A_T)Z^{(l-1)}W_T + b_T, Z_S' = \emptyset(A_S)Z^{(l-1)}W_S + b_S, \tag{15}$$

$$(Z_{ST}^{(l-1)})' = W_{Proj}((Z_T' \odot V_T)||(Z_S' \odot V_S)) + b_{Proj}, \tag{16}$$

where $W_{Q_t}, W_{K_t}, W_{Q_s}, W_{K_s} \in \mathbb{R}^{C_{in}\times C_d}$, $b_T, b_S \in \mathbb{R}^{C_d}$, $W_{Proj} \in \mathbb{R}^{C_d\times rC_{in}}$, $b_{Proj} \in \mathbb{R}^{rC_{in}}$, $W_{V_t}, W_{V_s}, W_T, W_S \in \mathbb{R}^{C_{in}\times C_d}$ are all learnable parameters for linear transformation. Since compressing the spatiotemporal dimensions can yield more satisfactory results, $C_d$ is generally smaller than $C_{in}$. Here, $A_T \in \mathbb{R}^{T\times T}$ and $A_S \in \mathbb{R}^{N\times N}$ represent the temporal dynamic attention matrix and the spatial dynamic attention matrix, respectively.

*(2) Gravity-informed ST-GC2former Block*

A well-documented challenge in deep transformer-based models is over-smoothing [28], where the attention matrix loses its ability to focus with increasing model depth—a problem that is particularly acute in spatial domains with numerous nodes. Fortunately, the gravity model is well-suited to address this issue, as its physically meaningful spatial interaction matrix can mitigate such uninformative redundancy. However, effectively integrating this physical constraint into node-level prediction tasks to alleviate the over-smoothing phenomenon remains a key challenge.

Inspired by RetNet [83], we integrate this physical constrain into the attention matrix using a simple yet effective Hadamard product $\odot$. This allows the gravity model to serve as a learnable distance decay pattern that re-weights the learned attention scores for node-oriented prediction tasks. Specifically, it can be represented as follows:

$$Z'_{S_g} = \emptyset\left(A_S \odot A_{ag}\right) Z^{(l-1)} W_S + b_S, \tag{17}$$

$$(Z_{GST}^{(l-1)})' = W_{Proj}((Z'_T \odot V_T) || \left(Z'_{S_g} \odot V_S\right)) + b_{Proj}, \tag{18}$$

where $\emptyset\left(A_S \odot A_{ag}\right)$ denotes the gravity attention matrix, And the rest is the same as the above ST-GC2former.

Similar to conventional transformer-based models, we also utilize a feed forward network (FFN) and layer normalization (LayerNorm) to achieve the high performance. However, instead of a standard FFN, a Gated Linear Unit Feed Forward Network (GLU-FFN) is employed to achieve the information filtering and improve gradient flow through a gating mechanism [84]. Specifically, the proposed GST-GC2former block can be represented as follows:

$$Z_{state}^{(l-1)} = (Z^{(l-1)})'_{GST} + Z^{(l-1)}, \tag{19}$$

$$Z^{(l)} = \mathcal{F}_{GFN}(Z_{state}^{(l-1)}) + Z_{state}^{(l-1)}, \tag{20}$$

where $\mathcal{F}_{GFN}(X) = \epsilon(W_1 X + b_1) \odot (W_2 X + b_2)$ denotes the GLU-FFN, $\epsilon(\cdot)$ is the GELU activation function, and $W_1, W_2 \in \mathbb{R}^{rC_{in} \times C_d}, b_1, b_2 \in \mathbb{R}^{C_d}$ are the learnable parameters in FFN. $Z_{state}^{(l-1)}$ denotes the hidden state of the $(l-1)$ layer of the GST-GC2former block.

## *E. Gravityformer*

Based on the above modules, our Gravityformer is proposed. Before entering the GST-GC2former, the embedding feature $Z$ is processed by the Hadamard mapper [85-87]. The primary motivation for this step is to address the feature sparsity and anomalous peaks inherent in human activity, which can be considered a form of noise that hinders effective learning. Since our human activity intensity data is derived from visits lasting longer than 30 minutes, many spatial units exhibit very low activity during non-peak hours, leading to a highly sparse input feature matrix. The Hadamard mapper acts as an efficient feature diffuser to mitigate this issue. By applying an orthogonal Hadamard transform, it redistributes the information from localized, high-magnitude peaks across the entire feature dimension. This transformation converts a sparse vector with sharp anomalies into a denser, more regularized representation. This pre-processing provides a more uniform input for the subsequent transformer layers, which stabilizes the attention mechanism and mass parameters of AdaGravity. Specifically, it employs a block diagonal matrix $H_p$ to distribute anomalous information to nearby tokens in the matrix:

$$H_p = \frac{1}{\sqrt{2}} \begin{bmatrix} H_{p-1} & H_{p-1} \\ H_{p-1} & -H_{p-1} \end{bmatrix} \tag{21}$$

$$Z^{(0)} = (Z H_p) H_p^{\intercal} (W_H H_p) H_p^{\intercal} \tag{22}$$

where $W_H \in \mathbb{R}^{C_{in} \times C_{in}}$ is a learnable projection parameter, and the initial value of $H_0$ is 1.

After passing through $L$ GST-GC2former layers, multiple residual skip connections are conducted to retain the hidden features of each layer. Then, these features are decoded through a simple MLP head. It can be expressed as follow:

$$Z^{(l)} = \left(W_{en}^{(l-1)} \mathcal{F}_{GST}\left(Z^{(l-1)}\right) + b_{en}^{(l-1)}\right) + Z^{(l-1)}, \tag{23}$$

$$\hat{Y} = W_{dec2} \phi(W_{dec1}(Z^{(L)}) + b_{dec1})) + b_{dec2}, \tag{24}$$

where the $\mathcal{F}_{GST}(\cdot)$ denotes the aforementioned GST-GC2former block, and $W_{en}^{(l-1)} \in \mathbb{R}^{C_d \times C_{skip}}, b_{en}^{(l-1)} \in \mathbb{R}^{C_{skip}}$ are the learnable parameters of the $(l-1)$ layer of the residual skip connection. And $Z^{(L)} \in \mathbb{R}^{N \times T \times C_{skip}}$ is the output of $L$ GST-GC2former, and $W_{dec1} \in \mathbb{R}^{C_{skip} \times C_d}$ , $b_{dec1} \in \mathbb{R}^{C_d}$ , $W_{dec2} \in \mathbb{R}^{C_d \times 1}, b_{Proj} \in \mathbb{R}^1$ are also the learnable parameters of decoder for final prediction.

Finally, the L1 loss is selected as our loss function for training our model:

$$\mathcal{L}_{L1}\left(Y, \hat{Y}\right) = \frac{1}{n} \sum_{i=1}^{n} \left|Y_i - \hat{Y}_i\right|, \tag{25}$$

where $Y \in \mathbb{R}^{N \times T \times 1}$ denotes the ground truth of human activity intensity and $\hat{Y} \in \mathbb{R}^{N \times T \times 1}$ dnotes the prediction of Gravityformer.

To provide a clear understanding of the training process for Gravityformer, the pseudocode is as follows:

**Table 1** The pseudocode of training Gravityformer.

**Algorithm** The training process of Gravityformer.

1: **Input:**
2: Historical human activity intensity data, inflow data and outflow data: $X_H = \{X_{t-q+1}, X_{t-q+2}, \ldots, X_t\}$, $X_{in} = \{X_{t-q+1}^{in}, X_{t-q+2}^{in}, \ldots, X_t^{in}\}$, $X_{out} = \{X_{t-q+1}^{out}, X_{t-q+2}^{out}, \ldots, X_t^{out}\}$,
3: Time-stamps data of time-of-day, day-of-week, and whether holiday: $S_H^s = \{S_{t-q+1}^s, S_{t-q+2}^s, \ldots, S_t^s\}$, $S_H^w = \{S_{t-q+1}^w, S_{t-q+2}^w, \ldots, S_t^w\}$, $S_H^h = \{S_{t-q+1}^h, S_{t-q+2}^h, \ldots, S_t^h\}$
4: The initialized spatiotemporal adaptive embedding: $\mathcal{F}_{stae}(\cdot)$
3: The initialized adaptive gravity model: $\mathcal{F}_{ag}(\cdot)$
4: The initialized ST-GC2former of Gravityformer: $\mathcal{F}_{ST}(\cdot)$
5: The initialized Hadamard mapper of Gravityformer: $\mathcal{F}_{hm}(\cdot)$
6: The initialized MLP Head of Gravityformer: $\mathcal{F}_{mlp}(\cdot)$
7: The initialized residual weights of Gravityformer: $\mathcal{F}_{res}(\cdot)$
8: Hyperparameter: Predicted snapshots: $p$, Historical snapshots: $q$, Learning rate $r$

9: Ground Truth: $Y_P = \{Y_{t+1}, Y_{t+2}, \ldots, Y_{t+p}\}$
10: **Output:**
11: Learned parameters of Gravityformer: $\Theta$
12: Prediction: $Y_{pre} = \{Y_{t+1}, Y_{t+2}, \ldots, Y_{t+p}\}$
13: **procedure** Gravityformer Training
14: represent the spatiotemporal feature $Z = \mathcal{F}_{stae}(X_H, S_H^s, S_H^w, S_H^h)$
15: transform the feature with Hadamard Mapper $Z^{(0)} = \mathcal{F}_{hm}(Z)$
16: **for** $l$ in $0{:}L$ **do**
17: compute the adaptive gravity matrix $A_{ag} = \mathcal{F}_{ag}(X_{in}, X_{out}, Z, Z^{(l-1)})$
18: update the feature with ST-GC2former $Z' = \mathcal{F}_{ST}(Z^{(l-1)}, A_{ag})$
19: residual learning $Z^{(l)} = \mathcal{F}_{res}(Z') + Z^{(l-1)}$
20: compute the prediction $Y_{pre} = \mathcal{F}_{mlp}(Z^{(L)})$
21: calculate the loss from prediction and ground truth $Loss = \mathcal{L}_{L1}(Y_{pre}, Y_P)$
22: updates the learnable parameters $\Theta$ based on their gradients and learning rate $r$
23: **end for**
24: **repeat**
25: Optimize the model parameters by Adam optimizer to minimize $Loss$
26: **until** convergence

## IV. Experiments and Results

### A. Datasets

To evaluate the effectiveness of the proposed model, large-scale human activity intensity data from six real-world metropolitan areas was utilized for experiments validation. Specifically, the datasets have obvious difference in spatial scale, including Atlanta, Boston, Chicago, Los Angeles, Minneapolis, and San Francisco (Fig. 7). To reduce unpredicting sparse activity in some areas, only the built-up areas of these cities were selected as the study regions.

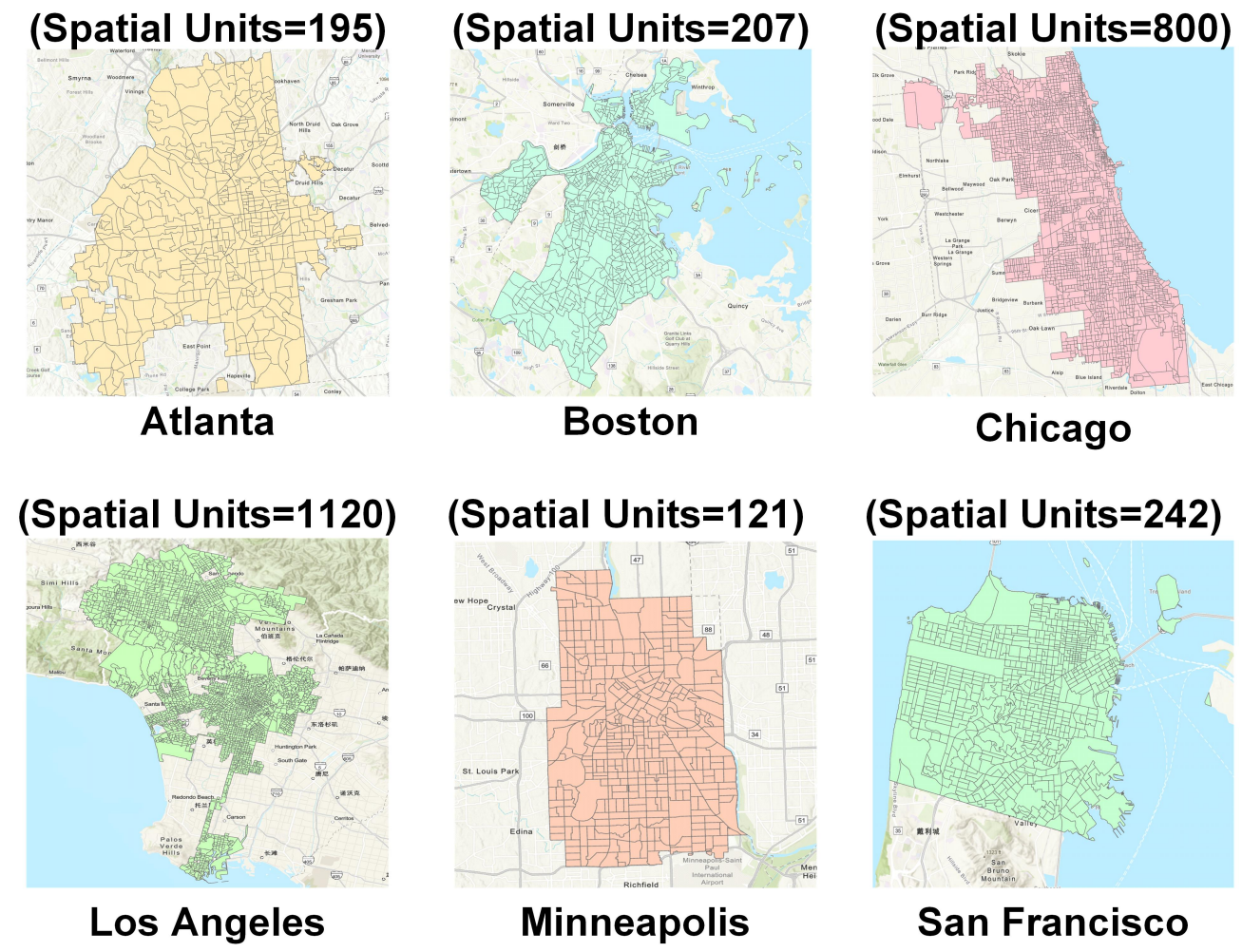

**Fig. 7** The study area consists of six metropolises with spatial units at the census tract level.

Within these study areas, a large volume of device-level cell phone data provided the foundation for our research. It was collected by PlaceIQ (https://www.placeiq.com/), containing on about hundreds of millions of anonymized visit and movement records per calendar day. These data have been randomly sampled at a rate of 5-10% for individual privacy protecting. To distinguish between activity and movement behaviors, the active visit data is filtered with a stay duration of more than 30 minutes to represent local human activity intensity. The start and end of a visit are defined as an inflow and an outflow, respectively. The total human activity intensity is the sum of inflows and outflows within 30 minutes. Data of those six datasets ranging from January 1, 2021, to January 31, 2022, were used in our experiments (i.e., January, April, July, November, December of 2021, and January of 2022). Then, these data were aggregated at the census tract level for further study and analysis.

After filtering out some anomalous records, 168 days and a total of 8064 snapshots were selected for each dataset, with each time step representing a half-hour interval. And the datasets were divided chronologically into training, validation, and test sets in a ratio of 7:1:2. To facilitate model training, Z-score normalization was applied to simplify the parameters learning of the model. All models used observations from the previous six hours to predict the next two hours.

### B. Benchmarks

The following time-series and spatiotemporal forecasting methods, ranging from classical to state-of-the-art, were selected as baselines for comparison with our Gravityformer.

It includes: 1) Classical Time-Series Forecasting Methods such as HA [88], FC-LSTM [89], TCN [90]; 2) Latest time-series forecasting methods like Dlinear [91] and iTransformer [92]; 3) Graph-based spatiotemporal forecasting methods, including STGCN [36], ASTGCN [43], Graph WaveNet [93], AGCRN [44], DSTAGNN [94], MegaCRN [95], ST-Wave [37], GDGCN [96], HyGCN [17], STID [52], HimNet [48], STG-Mamba [97]; and 4) Transformer-based spatiotemporal forecasting methods such as STTN [49], ASTGNN [38], BigST [54], STAEformer [39].

### C. Experiments Settings

The experiments were conducted using the PyTorch framework with an NVIDIA A800 graphics card. Specifically, we set the channels of GST-GC2former $C_{in}$ to 32, the squeezed transform channels $C_d$ to 8, the residual skip channels $C_{skip}$ to 256 and the number of GST-GC2former blocks $L$ to 6. The proposed model used the same hyperparameters with all datasets. In the training process, the Adam optimizer was selected for 200 epochs of model training with an initial learning rate of 0.002 and a weight decay of 0.0005. The all parameters of our model are initialized using the Xavier uniform method and the batch size was set to 16. According to the reference [39], the adaptive embedding parameters $C_t$, $C_a$ and $C_f$ are set to 24, 80 and 24, respectively. Then, the scaling parameters $\kappa$ and $C_e$ is set to 0.2 and 40, respectively, based on adaptive graph generation [45]. Additionally, since our models involve inflows and outflows, all models take inflows, outflows, and human activity intensity as inputs to ensure fairness. And all hyperparameters of those benchmarks were setting according to their original paper.

To assess the results, three widely used regression metrics are conducted to measure the performance of our model and benchmarks, i.e., root mean square error (RMSE), mean absolute error (MAE), and mean absolute percentage error (MAPE). For detail, the calculation formulas of those metrics are as follows:

$$RMSE = \sqrt{\frac{1}{n}\sum_{i=1}^{n}\left(Y - \hat{Y}\right)^{2}}, \tag{26}$$

$$MAE = \frac{1}{n}\sum_{i=1}^{n}\left|Y - \hat{Y}\right|, \tag{27}$$

$$MAPE = \frac{100\%}{n}\sum_{i=1}^{n}\left|\frac{Y-\hat{Y}}{Y}\right|, \tag{28}$$

*D. Results*

*(1) Overall Performance*

Compared with other models on six datasets, our Gravityformer outperformed all baselines in most cases (Table 2), including some state-of-the-art models, such as STAEformer, HimNet and STWave. Specifically, it demonstrated more excellent performance across all experiments, with the highest improvement of approximately 14% RMSE. A two-sided t-test confirms that our proposed model achieves a statistically significant improvement over the best-performing baseline ($p < 0.025$).

It can be observed that deep learning-based models generally achieved better results. The time-series forecasting models, such as FC-LSTM, TCN, Dlinear, and iTransformer, which consider only temporal variations and ignore spatial dependencies, showed limited performance on spatiotemporal forecasting task. In comparison, the ST-GNNs models performed better due to their ability to address spatial modeling challenges. Among these, STAEformer, as a representative state-of-the-art model, achieved good results across all datasets, confirming the significance of correctly representing potential spatial dependencies in the spatiotemporal context. Unfortunately, all these models are purely data-driven and overlook the spatial interactions underlying physical laws. This is why our proposed Gravityformer model achieved superior results in almost metrics.

Additionally, as the spatial scale of the datasets increased, we found that the differences among the baseline models gradually diminished. Contrastively, the proposed Gravityformer demonstrated more stable improvement under larger spatial scales, particularly in RMSE. To clearly illustrate the differences, we visualized the prediction and ground truth over three days in Minnesota using time-series histograms (Fig. 8). Each line represents dynamic changes in human activity at different regions, with darker colors indicating higher RMSE of models' prediction. It can be observed that the error coloration for Gravityformer is lighter, indicating lower prediction errors. It is evident that proposed model demonstrates significantly lower RMSE compared to other state-of-the-art models, particularly in regions with high human activity intensity. It indicates that incorporating physical laws into neural networks (i.e., the PIML) can help improve model stability and reduce the occurrence of significant errors.

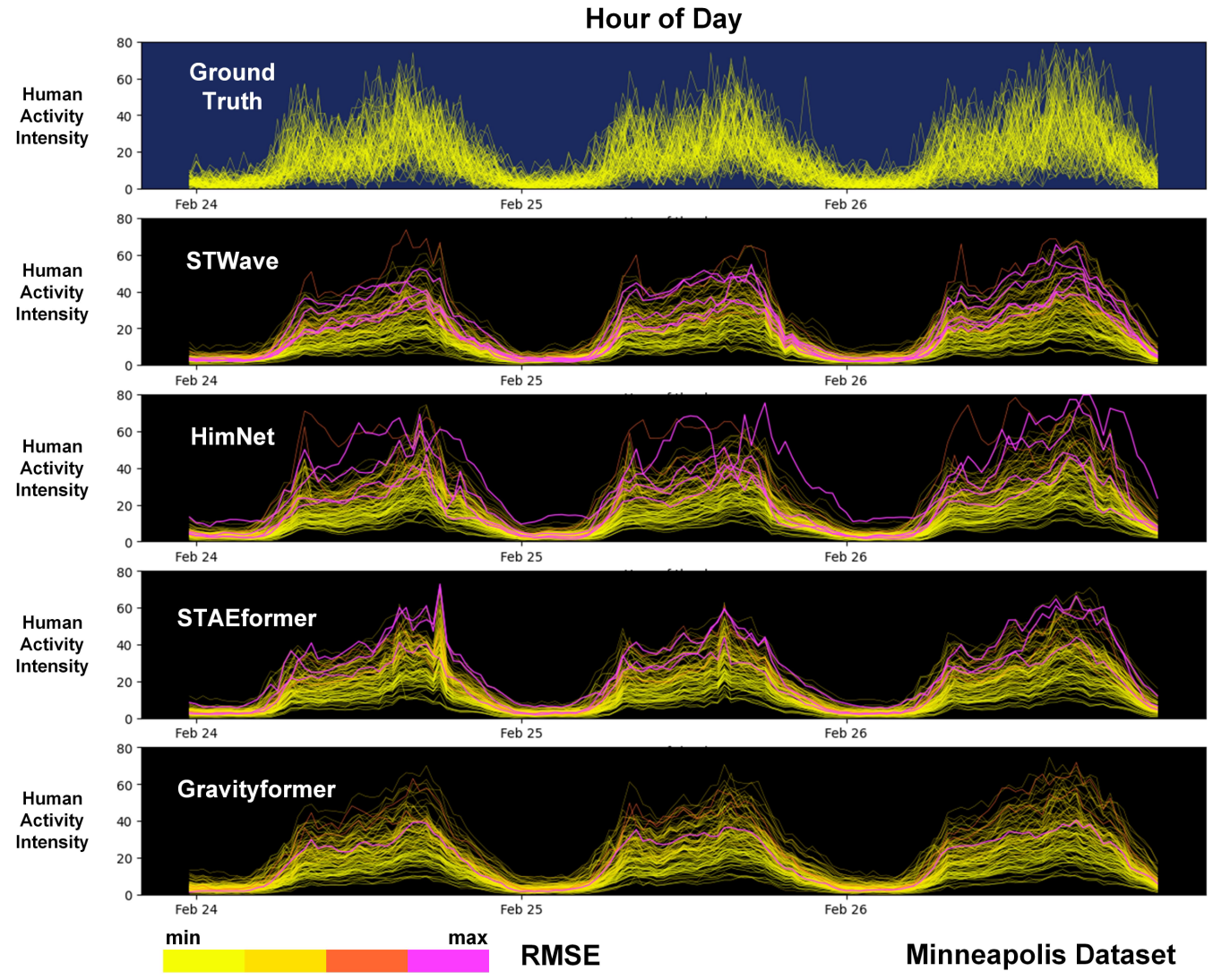

**Fig. 8** Time series histograms of ground truth and the prediction of different models in Minneapolis dataset from 24 Feb. 2021 to 26 Feb. 2021.

**Table 2** Performance of different models on six human activity datasets.

| Datasets \ Benchmarks (Journal/Conference) | Minneapolis | | | Atlanta | | | Boston | | |
|---|---|---|---|---|---|---|---|---|---|
| | RMSE | MAE | MAPE | RMSE | MAE | MAPE | RMSE | MAE | MAPE |
| HA | 52.83 | 19.19 | 131.04 | 66.92 | 25.66 | 137.42 | 39.76 | 19.05 | 139.80 |
| FC-LSTM | 40.89 | 11.76 | 63.36 | 48.16 | 14.26 | 60.30 | 25.63 | 11.22 | 61.99 |
| TCN | 36.07 | 11.52 | 62.66 | 43.93 | 13.94 | 60.64 | 24.58 | 11.07 | 66.86 |
| Dlinear (23' AAAI) | 53.56 | 18.58 | 58.35 | 56.50 | 19.09 | 68.31 | 29.08 | 12.68 | 71.27 |
| iTransformer (24' ICLR) | 42.38 | 10.49 | 48.15 | 46.83 | 13.45 | 54.11 | 22.62 | 9.64 | 52.71 |
| STGCN (18' IJCAI) | 37.46 | 9.98 | 48.18 | 41.98 | 11.75 | 51.31 | 21.76 | 9.34 | 51.96 |
| ASTGCN (19' AAAI) | 37.01 | 9.35 | 47.85 | 48.74 | 11.83 | 45.62 | 22.58 | 9.32 | 54.44 |
| Graph WaveNet (19' IJCAI) | 32.07 | 9.47 | 47.52 | 38.13 | 11.18 | 49.77 | 20.00 | 8.72 | 50.56 |
| STTN (20' arXiv) | 38.53 | 9.85 | 51.16 | 42.02 | 11.70 | 49.71 | 20.16 | 9.03 | 55.04 |
| AGCRN (20' NIPS) | 51.74 | 10.61 | 48.15 | 52.62 | 11.80 | 46.46 | 29.41 | 9.45 | 51.49 |
| ASTGNN (21' TKDE) | 43.04 | 9.70 | 47.78 | 44.67 | 10.94 | 44.57 | 19.82 | 8.46 | 51.57 |
| DSTAGNN (22' ICML) | 36.26 | 9.35 | 46.83 | 39.79 | 10.84 | 45.96 | 18.56 | 8.30 | 49.84 |
| STID (22' CIKM) | 31.84 | 9.66 | 47.90 | 37.93 | 11.38 | 48.56 | 20.73 | 9.13 | 54.10 |
| MegaCRN (23' AAAI) | 45.69 | 9.80 | 47.78 | 45.08 | 10.96 | 45.24 | 20.27 | 8.44 | 49.53 |
| GDGCN (23' INFFUS) | 29.43 | 9.21 | 50.36 | 38.04 | 11.38 | 50.10 | 19.39 | 8.78 | 50.69 |
| STWave (23' ICDE) | 29.63 | 9.15 | 46.77 | 38.80 | 10.48 | 46.33 | 18.45 | 8.45 | 51.82 |
| STAEformer (23' CIKM) | 38.31 | 9.00 | 45.11 | 41.61 | 10.43 | **42.71** | 18.34 | 8.17 | 48.13 |
| HyGCN (24' INFFUS) | 32.55 | 9.71 | 50.38 | 38.29 | 11.59 | 49.23 | 19.48 | 8.89 | 51.74 |
| BigST (24' VLDB) | 41.06 | 9.94 | 46.61 | 45.26 | 12.04 | 44.92 | 21.93 | 9.14 | **48.08** |
| STG-Mamba (24' arXiv) | 40.38 | 9.70 | 46.61 | 43.63 | 11.28 | 46.96 | 19.81 | 8.47 | 50.55 |
| HimNet (24' KDD) | 46.11 | 9.55 | 46.48 | 50.45 | 11.21 | 43.98 | 21.67 | 8.40 | 49.01 |
| **Gravityformer (Ours)** | **26.45** | **8.63** | **44.69** | **36.75** | **10.15** | 43.96 | **16.90** | **7.97** | 49.51 |

**Table 2** *Continuous.*

| Datasets \ Benchmarks (Journal/Conference) | San Francisco | | | Chicago | | | Los Angeles | | |
|---|---|---|---|---|---|---|---|---|---|
| | **RMSE** | **MAE** | **MAPE** | **RMSE** | **MAE** | **MAPE** | **RMSE** | **MAE** | **MAPE** |
| HA | 39.97 | 18.43 | 128.42 | 49.61 | 20.90 | 117.44 | 39.04 | 18.89 | 110.92 |
| FC-LSTM | 23.67 | 10.57 | 61.78 | 35.11 | 12.46 | 57.30 | 25.55 | 11.34 | 57.28 |
| TCN | 23.11 | 10.49 | 61.90 | 30.15 | 12.19 | 57.98 | 24.06 | 11.22 | 59.46 |
| Dlinear (23' AAAI) | 19.14 | 8.65 | 50.29 | 39.06 | 14.76 | 68.10 | 30.64 | 13.02 | 64.20 |
| iTransformer (24' ICLR) | 21.71 | 9.22 | 54.82 | 30.90 | 10.66 | 48.43 | 23.00 | 9.74 | 48.24 |
| STGCN (18' IJCAI) | 20.02 | 8.76 | 50.89 | 26.35 | 9.76 | 46.48 | 19.93 | 9.02 | 44.93 |
| ASTGCN (19' AAAI) | 23.75 | 8.89 | 48.44 | 27.02 | 9.67 | 45.61 | 39.51 | 10.43 | 46.63 |
| Graph WaveNet (19' IJCAI) | 18.38 | 8.40 | 50.80 | 26.56 | 9.62 | 46.90 | 19.41 | 8.91 | 47.11 |
| STTN (20' arXiv) | 20.19 | 8.72 | 49.85 | 28.47 | 9.88 | 45.08 | 21.30 | 9.35 | 47.84 |
| AGCRN (20' NIPS) | 28.67 | 8.70 | **46.18** | 26.57 | 9.64 | 47.26 | 19.48 | 8.97 | 47.75 |
| ASTGNN (21' TKDE) | 20.84 | 8.26 | 50.23 | 38.62 | 9.85 | 42.78 | 33.87 | 9.00 | 44.68 |
| DSTAGNN (22' ICML) | 17.95 | 7.88 | 47.18 | 29.21 | 9.24 | 43.92 | 19.06 | 8.45 | 44.78 |
| STID (22' CIKM) | 18.76 | 8.63 | 50.76 | 26.26 | 9.95 | 47.38 | 19.57 | 9.20 | 46.25 |
| MegaCRN (23' AAAI) | 21.07 | 8.04 | 47.07 | 34.20 | 9.47 | 44.13 | 36.37 | 9.03 | 43.26 |
| GDGCN (23' INFFUS) | 18.11 | 8.46 | 50.44 | 25.06 | 9.49 | 46.13 | 18.65 | 9.02 | 47.28 |
| STWave (23' ICDE) | 17.63 | 7.98 | 47.45 | 27.66 | 9.72 | 48.10 | 19.79 | 9.18 | 48.23 |
| STAEformer (23' CIKM) | 17.94 | 7.73 | 46.63 | 27.88 | 9.00 | 42.82 | 20.65 | 8.32 | **41.52** |
| HyGCN (24' INFFUS) | 18.94 | 8.67 | 50.26 | 25.31 | 9.79 | 45.28 | 19.20 | 9.15 | 46.65 |
| BigST (24' VLDB) | 19.85 | 8.41 | 49.92 | 29.14 | 10.01 | 44.74 | 22.27 | 8.92 | 47.19 |
| STG-Mamba (24' arXiv) | 19.06 | 8.08 | 46.40 | 31.90 | 9.45 | 43.63 | 19.90 | 8.70 | 44.35 |
| HimNet (24' KDD) | 21.37 | 7.89 | 46.71 | 42.84 | 9.50 | 43.40 | 36.05 | 8.76 | 42.95 |
| **Gravityformer (Ours)** | **16.13** | **7.57** | 46.82 | **23.76** | **8.66** | **42.74** | **16.03** | **8.01** | 42.92 |

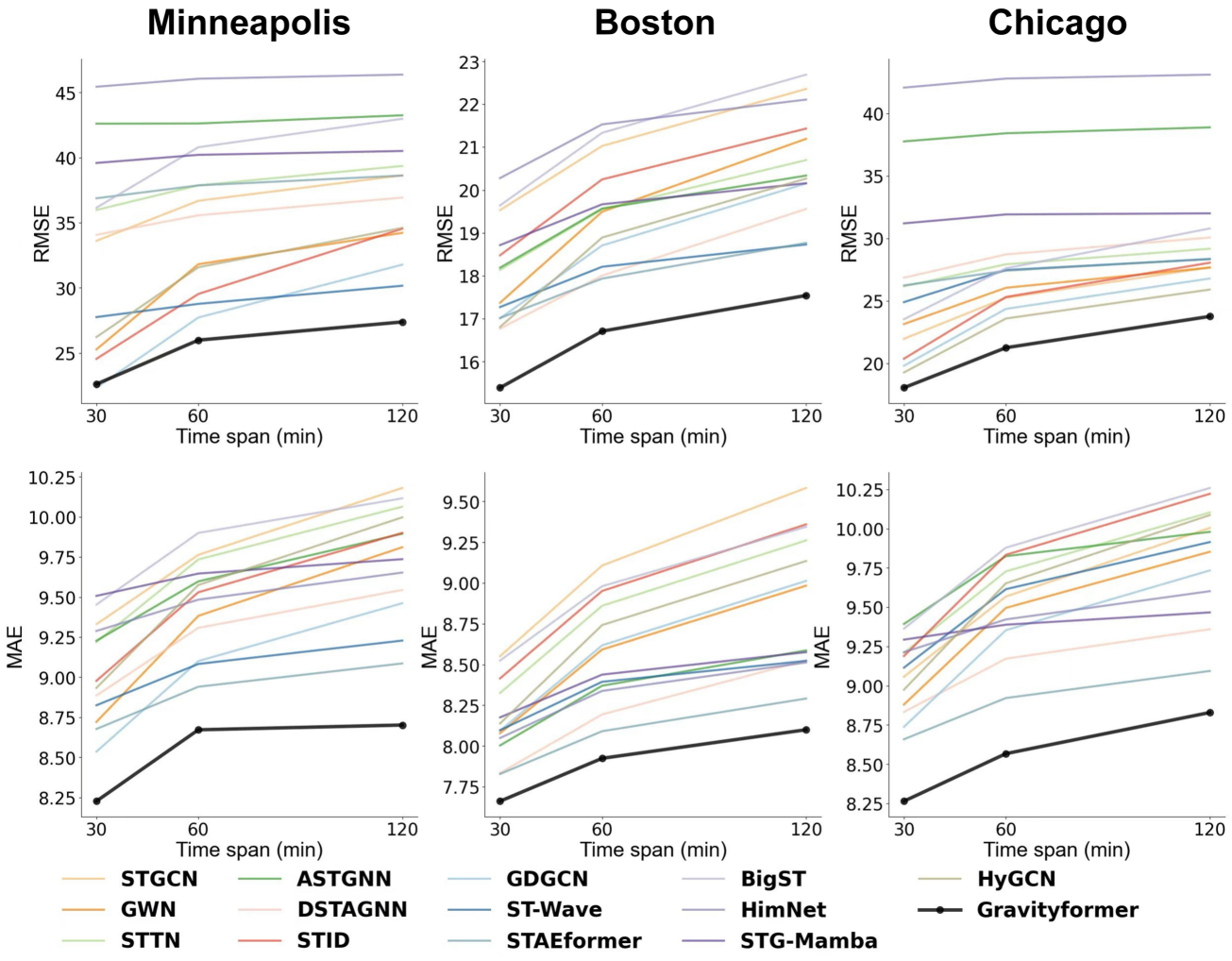

**Fig. 9** Multi-step prediction performance of different models.

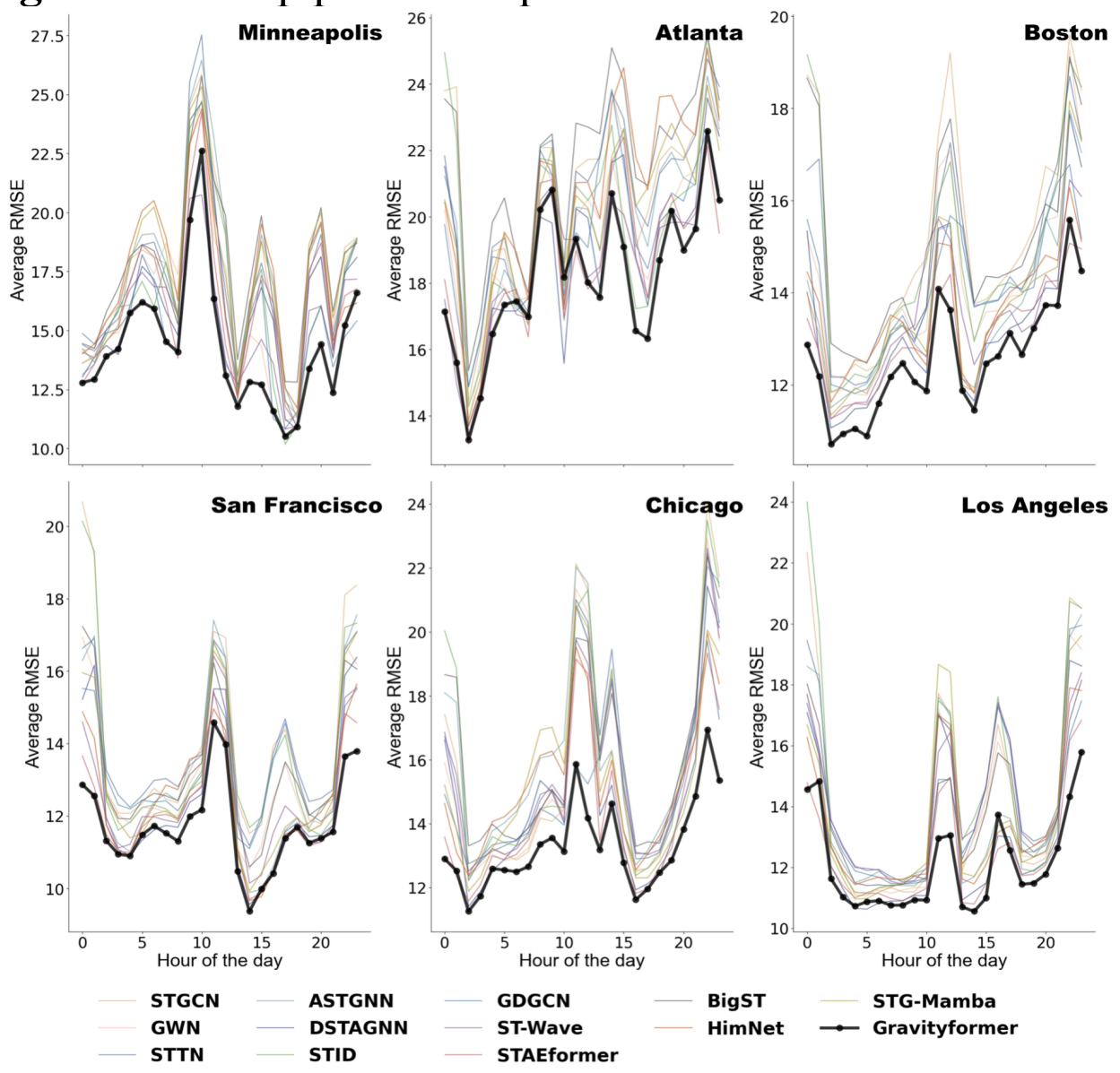

**Fig. 10** Hourly temporal average RMSE of different models on different datasets.

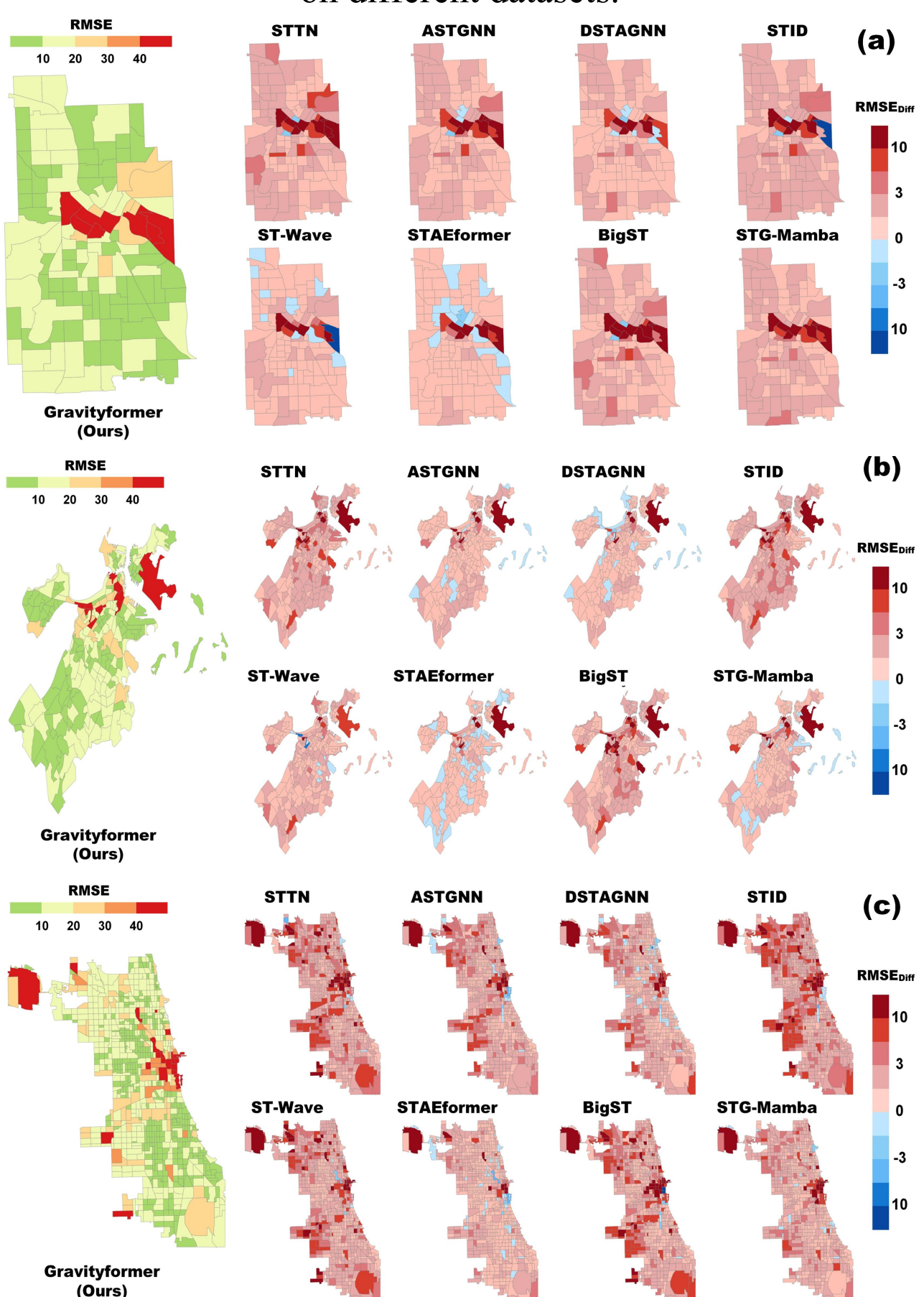

**Fig. 11** The spatial RMSE comparison of different models on (a) Minneapolis dataset (small dataset), (b) Boston dataset (median dataset) and (c) Chicago dataset (large dataset).

*(2) Temporal and Spatial Error Distributions*

To further demonstrate the advantages of our model compared to other models, we visualized and analyzed the errors distribution from both temporal and spatial dimensions.

On one hand, the average hourly RMSE was visualized in Fig. 10. Compared to other models, our Gravityformer consistently exhibited lower RMSE results throughout the day, especially during the evening peak hours from 17:00 to 21:00. This phenomenon indicates that the proposed model can reduce larger temporal errors and provide greater stability by leveraging the physical laws of human movement. This stability also offers a more reliable foundation for location-based services and geographic down-streamed tasks. On the other hand, we also selected three datasets—Minneapolis, Boston, and Chicago—representing small, medium, and large urban areas, respectively, for further spatial analysis (Fig. 19). Specifically, a new metric $RMSE_{Diff}$ was introduced to highlight the performance gap, which is the difference between the average RMSE of baselines and Gravityformer, while Gavityformer is shown in RMSE.

The results demonstrated that our Gravityformer achieved lower spatial errors across all datasets, with a particularly reduction in the Chicago. This finding further corroborates that the lack of physical laws makes prediction more challenging, especially in large-scale human activity datasets.

*(3) Error Analysis on Weekdays and Weekends*

We analyzed the prediction performance on distinct weekday versus weekend patterns across three datasets (Fig. 12). While Gravityformer outperforms baselines in both cases, the improvement (particularly in RMSE) is more pronounced on weekdays. We attribute this to the greater regularity of weekday activities (e.g., work schedules) compared to the more stochastic nature of weekends. This suggests our physics constraints are most effective in scenarios with stable, predictable patterns.

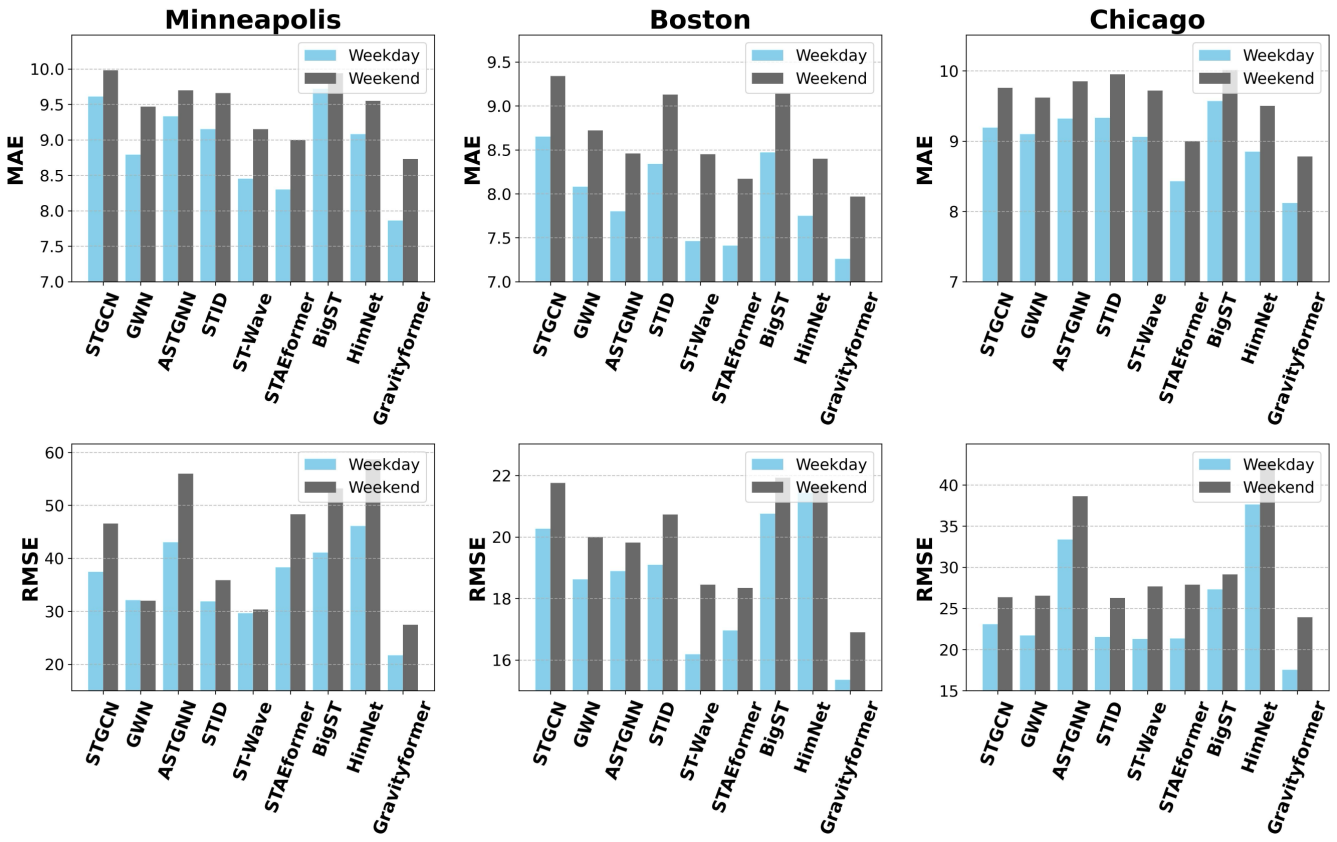

**Fig. 12** Comparative analysis of prediction on weekdays and weekends.

## V. Discussion

### *A. Ablation study*

To understand the impact of each component, we conducted an ablation study with comparative analysis. Seven variants were designed as baselines to discuss their significance. To ensure fairness, all variants were trained and tested with consistent parameters. Specifically, the definitions of the variants are as follows:

- w/o AdaGravity: This variant completely removes the AdaGravity component.
- w/o Adaptive distance scaling: This variant excludes the adaptive distance scaling from the AdaGravity component.
- w/o Inflow and outflow: This variant excludes the inflow and outflow features of the mass parameters in AdaGravity.
- w/o Distance matrix: This variant replaces the distance matrix in AdaGravity to an adaptive adjacency matrix.
- w/o Hadamard mapper: This variant excludes the hadamard mapper module.
- w/ Dynamic Parameters of AdaGravity: This variant achieves the dynamic learning for the key parameters of AdaGravity (the mass scaling factors $\alpha_1$ and $\alpha_2$, and the distance decay factor $\beta$) from feature $Z^{(l)}$.
- w/ Learnable Pooling for Mass Parameters: This variant replaces the simple average pooling with a learnable MLP to derive the mass parameters.
- w/o Conv2fomer structure: This variant lacks the dynamic spatial and temporal adjustment branches $V_S$ and $V_T$.
- w/o Parallel spatiotemporal structure: This variant removes the parallel spatiotemporal structure, replacing it with three layers of sequential temporal GC2former and three layers of spatial GC2former.

Given the performance of the variants (Table 3), several key observations can be made:

- AdaGravity contributes the most significant overall improvement, regardless of the dataset scale. Within the AdaGravity, the absence of inflow and outflow features (w/o Inflow and outflow) is relatively more critical, as these features help AdaGravity achieve dynamic temporal modeling. Although the adaptive adjacency matrix is beneficial, the results from the variant without distance matrix (w/ Distance matrix) indicate that the distance decay assumption is necessary for human activity intensity prediction. Then, the adaptive distance scaling (w/o Adaptive distance scaling) is effective, but not as critical as such components. Furthermore, we also investigated more complex designs for the AdaGravity module, such as implementing learnable pooling for the mass parameters or dynamic parameters for scaling factors. However, these attempts did not lead to consistent performance improvements. Instead, the increased complexity introduced difficulties in achieving convergence. This also confirms that the proposed model is simple and effective.
- Coupled spatiotemporal learning also provides performance improvements. Without the dynamic adjustment branches between $V_S$ and $V_T$ , the effectiveness of the coupled spatiotemporal learning diminishes, indicating their importance. Furthermore, replacing the parallel structure with a vertical structure revealed that dynamic adjustment can maximize the effectiveness of coupled spatiotemporal learning.

- The Hadamard mapper effectively suppresses noise. Since the predicted data only involves visits longer than 30 minutes, it would appear some spatial and temporal sparsity at nonpeak time. It can be mitigated by the Hadamard mapper, thereby improving the model performance.

**Table 3** Ablation experiments of Gravityformer on different datasets.

| **Ablation Study** | **Minneapolis** | | | **Boston** | | | **Chicago** | | |
|---|---|---|---|---|---|---|---|---|---|
| | **RMSE** | **MAE** | **MAPE** | **RMSE** | **MAE** | **MAPE** | **RMSE** | **MAE** | **MAPE** |
| w/o AdaGravity | 29.98 | 8.77 | 46.41 | 18.45 | 8.14 | 51.17 | 27.43 | 8.93 | 45.04 |
| w/o Adaptive distance scaling | 28.16 | 8.72 | 47.42 | 17.50 | 7.99 | 50.09 | 24.60 | 8.72 | 46.77 |
| w/o Inflow and outflow | 29.11 | 8.71 | 45.98 | 17.94 | 8.05 | 50.82 | 25.42 | 8.81 | 45.63 |
| w/ Distance matrix | 29.23 | 8.68 | 45.88 | 18.03 | 8.05 | 50.49 | 26.27 | 8.74 | 46.33 |
| w/ Dynamic Parameters of AdaGravity | 30.50 | 8.70 | 47.62 | **16.68** | **7.97** | 53.37 | 23.68 | 8.69 | 45.88 |
| w/ Learnable Pooling for Mass Parameters | 32.93 | 8.97 | 51.69 | 17.78 | 8.09 | 52.76 | **23.67** | 9.14 | 51.70 |
| w/o Hadamard mapper | 29.05 | 8.74 | 45.85 | 17.77 | 8.07 | 50.76 | 26.04 | 8.81 | 45.23 |
| w/o Conv2fomer structure | 27.52 | 8.65 | 46.94 | 17.24 | 7.99 | 50.12 | 24.54 | 8.72 | 44.12 |
| w/o Parallel spatiotemporal structure | 26.74 | 8.66 | 44.98 | 17.12 | 8.02 | **49.42** | 24.01 | 8.70 | 42.87 |
| **Gravityformer (Ours)** | **26.45** | **8.63** | **44.69** | 16.90 | **7.97** | 49.51 | 23.76 | **8.66** | **42.74** |

### *B. Spatially-explicitly analysis and interpretation*

To further validate the significance and interpretability of the proposed AdaGravity model, a series of quantitative and qualitative analyses were conducted. Specifically, we randomly selected an instance from the Gravityformer in the last layer of the GST-GC2former in Minneapolis to analyze the learned attention matrix, gravity attention matrix, and distance matrix through spatial explicit analysis.

Firstly, three original matrices were visualized for observing their patterns, as shown in Fig. 13(a). Specifically, the attention matrix exhibits values close to zero in most positions, which is the over-smooth phenomenon abovementioned by the Transformer. Contrastively, the gravity attention matrix does not exhibit such phenomenon, showing distinct differences across different positions. Moreover, it not only shares the same distribution patterns with the distance matrix, but also enhance key elements flexibly.

Then, a quantitative analysis of the three matrices was further conducted for their value distribution differences, as shown in Fig. 13(b). The values of the transformer attention are also primarily distributed at lower value positions. And the values of the gravity attention exhibit a clear distance decay effect. Compared with the distance matrix, the gravity attention matrix shows more diverse results, benefiting from the mass parameters. Specifically, the values at closer distances may not be as strongly correlated, while those at greater distances may not be as weakly correlated. This is likely due to the scaling of the distance and mass expansion coefficients. Additionally, we found that the gravity attention matrix also has a part of distribution around zero. This is because some human activity may be sparse, which could introduce unnecessary redundancy into the process of neighborhood feature aggregation. The gravity attention matrix can filter them well compared with others.

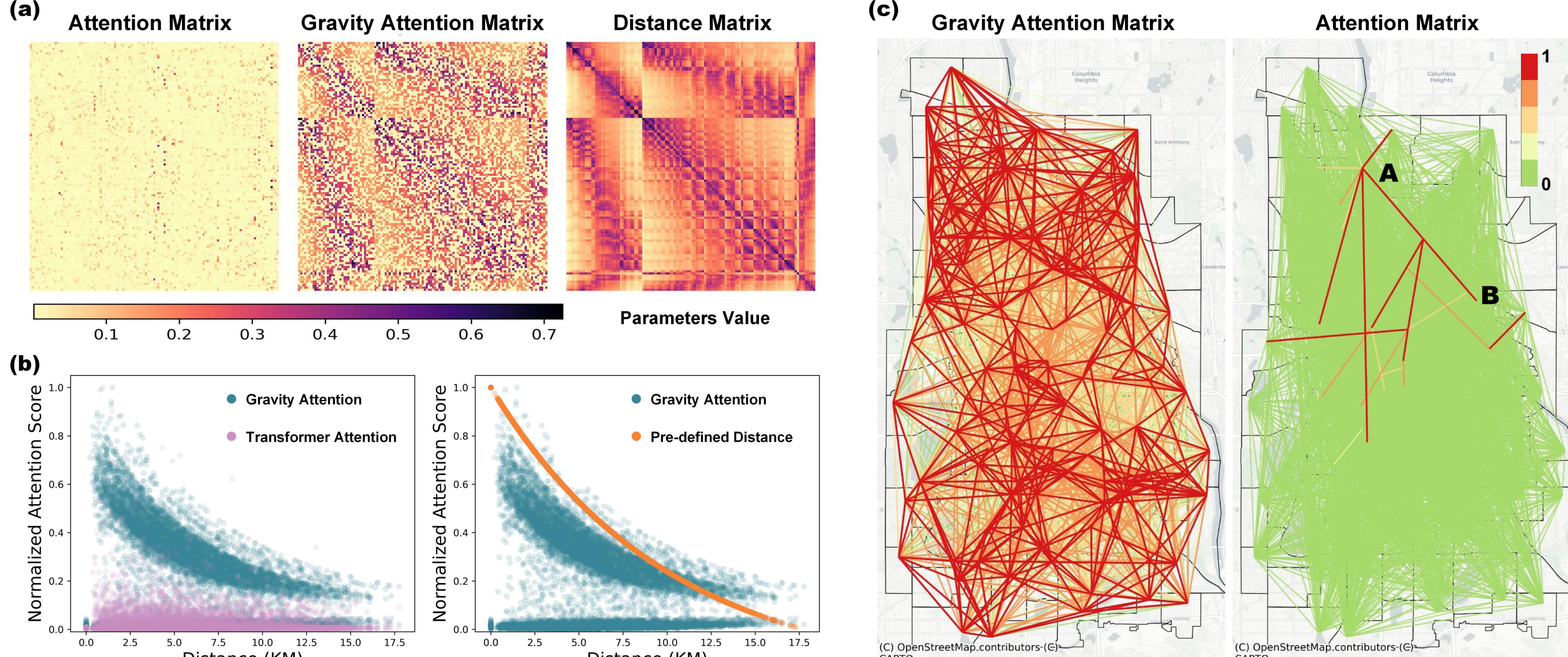

**Fig. 13** Spatially-explicitly analysis of learned matrix at final GST-GC2former block (a) The difference among attention matrix, gravity attention matrix and distance matrix of Minneapolis Dataset at $L = 6$ GST-GC2former Block. (b) the value distribution of such matrices. (c) The flow map of learned gravity attention matrix and attention matrix.

**Table 4** Configuration analysis of Gravityformer on different datasets.

| Hyperparameters | | Minneapolis | | | Boston | | | Chicago | | |
|---|---|---|---|---|---|---|---|---|---|---|
| | | RMSE | MAE | MAPE | RMSE | MAE | MAPE | RMSE | MAE | MAPE |
| $C_{in}$ | 16 | 26.88 | 8.69 | 45.55 | 17.32 | 8.08 | 49.89 | 24.64 | 8.78 | 43.22 |
| | 32 | **26.45** | 8.63 | **44.69** | **16.90** | **7.97** | 49.51 | 23.76 | 8.66 | **42.74** |
| | 64 | 26.58 | **8.61** | 45.12 | 16.99 | 8.02 | **49.24** | **23.67** | **8.64** | 44.56 |
| | 128 | 26.76 | 8.65 | 45.88 | 17.06 | 8.06 | 50.03 | 23.82 | 8.69 | 44.34 |
| $C_d$ | 4 | 27.22 | 8.73 | 45.89 | 17.45 | 8.04 | 50.30 | 25.24 | 8.80 | 44.10 |
| | 8 | **26.45** | 8.63 | 44.69 | **16.90** | 7.97 | **49.51** | **23.76** | **8.66** | 42.74 |
| | 16 | 26.80 | **8.62** | **44.62** | 16.94 | **7.94** | 49.63 | 23.81 | 8.68 | **42.62** |
| | 32 | 26.52 | 8.70 | 45.63 | 16.98 | 7.99 | 49.84 | 23.89 | 8.70 | 43.04 |
| $L$ | 4 | 27.12 | 8.80 | 45.34 | 17.45 | 8.10 | 50.83 | 24.88 | 8.76 | 44.76 |
| | 5 | 26.63 | **8.61** | 45.01 | 17.04 | 8.01 | **49.23** | 24.20 | **8.63** | 43.19 |
| | 6 | **26.45** | 8.63 | **44.69** | **16.90** | **7.97** | 49.51 | **23.76** | 8.66 | **42.74** |
| | 7 | 26.91 | 8.72 | 44.93 | 16.96 | 7.98 | 49.55 | 23.95 | 8.71 | 42.87 |

**Table 5** Efficiency comparison on the different dataset.

| Dataset / Benchmarks | Minneapolis | | | Boston | | | Chicago | | |
|---|---|---|---|---|---|---|---|---|---|
| | Params. (M) | Train (s/epoch) | Inference (s) | Params. (M) | Train (s/epoch) | Inference (s) | Params. (M) | Train (s/epoch) | Inference (s) |
| BigST | **0.06** | **2.94** | **0.26** | **0.07** | **3.50** | **0.35** | **0.09** | **4.37** | **0.54** |
| STG-Mamba | 0.56 | 6.01 | 0.42 | 1.60 | 6.54 | 0.51 | 23.55 | 11.56 | 0.77 |
| STWave | 0.85 | 8.56 | 1.10 | 0.85 | 14.53 | 1.93 | 0.85 | 48.68 | 6.86 |
| HimNet | 1.21 | 8.29 | 0.90 | 1.21 | 10.89 | 1.52 | 1.22 | 49.31 | 9.03 |
| STTN | 0.30 | 5.42 | 0.61 | 0.30 | 12.30 | 1.34 | 0.30 | 81.10 | 7.83 |
| STAEformer | 1.16 | 8.95 | 0.87 | 1.24 | 18.68 | 1.79 | 1.81 | 108.25 | 10.06 |
| ASTGNN | 0.45 | 10.00 | 0.66 | 0.46 | 19.98 | 1.37 | 0.53 | 94.17 | 7.10 |
| DSTAGNN | 1.75 | 13.04 | 0.98 | 2.29 | 24.66 | 1.75 | 9.23 | 118.36 | 12.43 |
| **Gravityformer (Ours)** | 0.32 | 7.77 | 0.60 | 0.56 | 16.01 | 1.23 | 4.59 | 79.92 | 6.81 |

Finally, the differences between the gravity attention matrix and the attention matrix are further discussed through the flow maps, as shown in Fig. 13(c). The nodes of the transformer attention show very few effective connections, while the nodes of the gravity attention show connections that are larger in closer area. These two relationships can be reasonably explained by the first and third laws of geography [98, 99]. On the one hand, the first law of geography mainly states that near things are more related to each other than distant things. This expected distance decay effect in geography has also been significantly learned by our model. The law of universal gravitation accounts for the significance of locations beyond the distance decay effect, suggesting that spatial correlations should be appropriately scaled and weighted. In our work, this is demonstrated by the fact that while most nodes exhibit the distance decay effect, some adjacent areas may display less pronounced connections. On the other hand, the third law of geography mainly states that the more similar the geographic configurations are, the more similar the target attributes will be. This situation also appears in the attention matrix of the transformer. For example, the place A and the place B are both residential areas, they show strong connections. Unfortunately, due to the excessive sparsity of transformer attention, the third law of geography cannot be fully explained across all regions.

### *C. Configuration analysis*

To further investigate the influence of hyperparameter settings, the experiments of different configurations are conducted on partial datasets. Specifically, the channels $C_{in}$ and squeezed transform channels $C_d$ of GST-GC2former, and number of GST-GC2former blocks $L$ are discussed, as shown in Table 4.

Regardless of whether the dataset is large-scale or small-scale, setting hyperparameters too high or too low would also lead to performance reduce. Small parameters may result in underfitting, while large parameters may cause convergence issues during training. Based on these those experiments and results, our proposed model employs the best parameter settings to balance these concerns.

### *D. Computational efficiency analysis*

To assess the computational efficiency of the proposed model, we selected three datasets of varying spatial scales and compared the numbers of parameter (Params.), training time (Train) and inference time (Inference) with some advanced baselines (Table 5). Due to the less complex model design, BigST and STG-Mamba exhibited the most efficient computation. The computational efficiency is the main advantage of MLP-based and Mamba-based models compared to Transformer-based models [100]. Conversely, Transformer-based models mainly focus on greater scalability and stronger performance [101], which is evident not only in time-series forecasting but also in other popular domains, such as foundational models in natural language processing and computer vision [102].

Compared to other Transformer-based models, the proposed model demonstrated relatively high efficiency due to the low hidden units, especially when compared to ASTGNN and STAEformer. Additionally, the time interval between snapshots in human activity intensity tasks is generally larger than in real-time traffic flow prediction [16]. It provides more opportunities for models with higher complexity and better performance. Considering the performance improvements (Table 2), task demands and unique spatially-explicitly interpretability, the computational cost of Gravityformer is moderate and acceptable.

*E. Zero-shot cross-region inference*

To evaluate the generalization capabilities of our model, we assess its transferability via zero-shot cross-region inference (Table 6). Two challenging transfer scenarios were designed: from a small-scale to a large-scale dataset (Minneapolis to San Francisco) and from a large-scale to a small-scale dataset (Boston to Minneapolis). To ensure a fair comparison, we standardized the batch size across all baseline models and retrained them accordingly. And any model components with learnable parameters whose dimensions are dependent on the number of spatial nodes were removed, as they are inherently non-transferable in a zero-shot setting. This included adaptive embeddings (STID, STAEformer, BigST, and our Gravityformer), adaptive adjacency graph (Graph WaveNet and our Gravityformer), and any node-dependent bias terms within the attention mechanisms of Transformer-based models.

**Table 6** Zero-shot cross-scale regional transfer inference.

| Transfer Datasets \ Benchmarks | Minneapolis to San Francisco | | | Boston to Minneapolis | | |
|---|---|---|---|---|---|---|
| | **MAE** | **RMSE** | **MAPE** | **MAE** | **RMSE** | **MAPE** |
| STGCN | 11.58 | 27.22 | 69.34 | 12.10 | 45.60 | 56.59 |
| Graph WaveNet | 14.54 | 34.88 | 80.73 | 13.55 | 45.11 | 58.59 |
| STID | 11.50 | <u>26.59</u> | **62.16** | <u>11.53</u> | <u>43.74</u> | 62.40 |
| BigST | 11.70 | 26.62 | 67.32 | **11.51** | 44.10 | 59.22 |
| STAEformer | <u>11.25</u> | 28.08 | 62.85 | 12.14 | 44.53 | **49.04** |
| **Gravityformer w/o AdaGravity (Ours)** | 11.26 | 28.19 | 62.35 | 12.01 | 44.54 | <u>52.06</u> |
| **Gravityformer (Ours)** | **11.07** | **26.51** | 62.83 | 11.67 | **43.05** | 53.80 |

**Table 7** Zero-shot Transfer analysis for AdaGravity Parameters.

| **Datasets \ Transfer types** | | **Evaluation Metrics** | | | **Key Parameters of AdaGravity** | | |
|---|---|---|---|---|---|---|---|
| | | **MAE** | **RMSE** | **MAPE** | $\sigma(\alpha_1)$ | $\sigma(\alpha_2)$ | $\sigma(\beta)$ |
| **Minneapolis to San Francisco** | w/o Transfer | **8.70** | **18.49** | **52.46** | Minneapolis: {0.54, 1.24, 1.27, 0.69, 0.70, 0.69} => San Francisco: {0.54, 1.24, 1.26, 0.69, 0.69, 0.69} | Minneapolis: {0.54, 1.38, 1.43, 0.70, 0.69, 0.69} => San Francisco: {0.54, 1.39, 1.42, 0.69, 0.69, 0.69} | Minneapolis: {0.77, 1.35, 1.28, 1.38, 1.39, 1.38} => San Francisco: {0.77, 1.34, 1.28, 1.39, 1.39, 1.39} |
| | Only Transfer AdaGravity | **8.74** | **19.05** | **53.62** | | | |
| | All Parameters Transfer | 11.07 | 26.51 | 62.83 | | | |
| | w/o AdaGravity | 11.26 | 28.19 | 62.35 | | | |
| **Boston to Minneapolis** | w/o Transfer | **9.30** | **29.69** | **49.28** | Boston: {0.64, 1.43, 1.10, 0.68, 0.69, 0.69} => Minneapolis: {0.54, 1.24, 1.27, 0.69, 0.70, 0.69} | Boston: {0.63, 1.42, 1.53, 0.69, 0.69, 0.70} => Minneapolis: {0.54, 1.38, 1.43, 0.70, 0.69, 0.69} | Boston: {1.44, 1.59, 1.09, 1.39, 1.39, 1.40} => Minneapolis: {0.77, 1.35, 1.28, 1.38, 1.39, 1.38} |
| | Only Transfer AdaGravity | **9.62** | **33.09** | **54.81** | | | |
| | All Parameters Transfer | 11.67 | 43.05 | 53.80 | | | |
| | w/o AdaGravity | 12.01 | 44.54 | 52.06 | | | |

Our analysis yields two key insights. First, in a comparative evaluation against several well-known baselines, Gravityformer exhibits competitive zero-shot performance. This indicates that the physics-informed constraints imposed by the AdaGravity module enhance the generalization capabilities. However, under zero-shot transfer scenarios, the performance differences among these models are modest. This finding indicates that robust zero-shot inference may rely more on transfer learning techniques rather than architectural improvements to the model backbone.

Furthermore, we also investigated the transferability of the AdaGravity module by analyzing its key parameters across datasets (Table 7). The core of the gravity model is defined by three critical parameters: the mass scaling factors, $\sigma(\alpha_1)$ and $\sigma(\alpha_2)$, and the distance decay factor, $\sigma(\beta)$. We designed a variant, "Only Transfer AdaGravity", which migrates only these three learned parameters to the new region. It can be observed that its performance was only marginally lower than model training on target dataset, highlighting the crucial role of these few parameters. Upon visualizing these key parameters from each AdaGravity layer, we found them to be similar across different datasets, especially in the Minneapolis-to-San-Francisco transfer. This stability implies that AdaGravity successfully learns a universal pattern of spatial interaction, which is key to enhance generalization of our model.

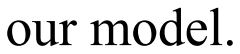

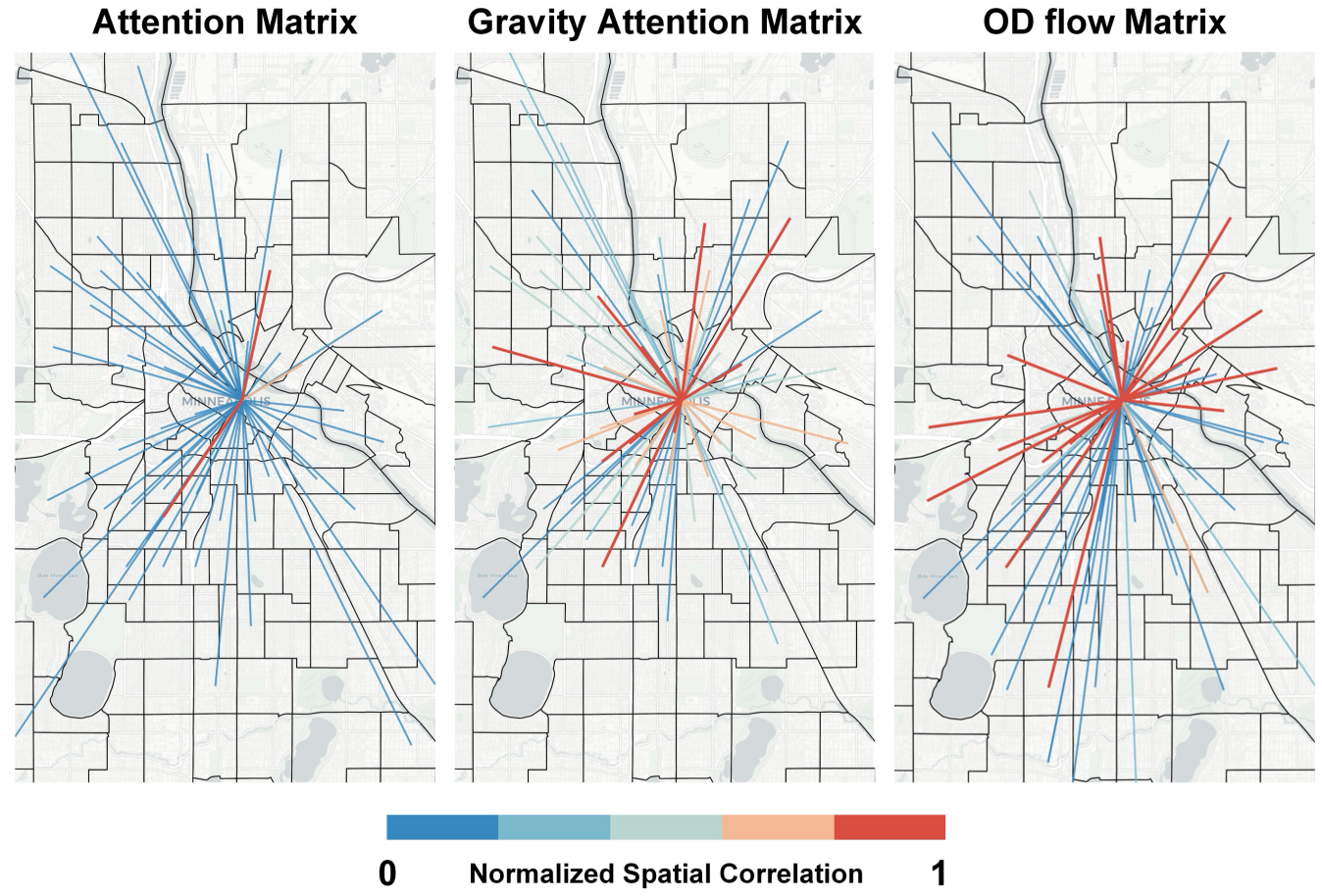

**Fig. 14** Comparison of gravity attention, standard attention, and real-world OD flow in special Area.

*F. Spatial analysis of special cases*

To further validate the rationale of our model, we conducted a case study in a special area: downtown Minneapolis, which is a major commercial hub and shopping center known for attracting visitors from a wide geographic area. We compared the spatial patterns learned by three matrices (Fig. 14): (1) the standard attention from the final layer, (2) our gravity attention from the final layer, and (3) the real OD flow. It can be observed that the learned gravity attention is more similar to the real OD matrix than the standard attention is. This implies that even in such special areas, Gravityformer enables to capture a pattern that more closely resembles real-world spatial interactions, which in turn confirms its interpretability again.

## VI. Conclusion

In this study, a gravity-informed spatiotemporal transformer, namely Gravityformer, was proposed for human activity intensity prediction. Inspired by PIML, the proposed model is designed for modeling the universal gravitation of human activity to not only provide the physics knowledge to those data-driven spatial learning module, but also reduce the over-smoothing phenomenon of transformer attention. Additionally, a parallel ST-GC2former module was proposed for balancing spatial and temporal coupled learning. Through systematic experiments, our Gravityformer outperforms state-of-the-art spatiotemporal forecasting model on human activity intensity prediction in the six real-world metropolitan datasets of human activity intensity. The spatial interpretability advantage of proposed model is also demonstrated through quantitative and qualitative analysis of learned spatial knowledge, aligning with the first and third laws of geography. Our finding demonstrates that such physics-informed deep learning enlightens a brighter future for spatially-explicit modeling in extensive applications of urban computing.

However, Gravityformer also have some limitations. While avoiding the need of fine-grained OD data, AdaGravity module relies on aggregated inflow and outflow data to learn mass parameters differentially. This requirement poses a data collection challenge, as such multi-attribute human activity datasets are often unavailable. Additionally, the zero-shot inference of our model is outstanding, as transfer learning were not a focus of this work. Future works could explore contrastive learning to alleviate data demand, and employ lifelong learning strategy or large language model to enhance ability of zero-shot inference.

## Acknowledgement

This work was supported by the High-performance Computing Platform of Peking University (hpc2306190166). This work was also partially funded by CTS Scholars Program, Center for Transportation Studies, University of Minnesota, and DSI Medium/Large Seed Grant, Data Science Initiatives, University of Minnesota. Additionally, the authors would like to thank Dr. Jing Li, Dr. Hao Guo and the anonymous referees for their insightful comments.

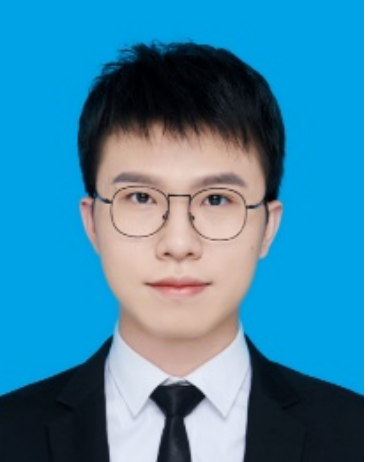

**Yi Wang** (Student Member, IEEE) is currently pursuing the Ph.D. degree at School of Earth and Space Sciences, Peking University. His main research interests include geospatial artificial intelligence, spatio-temporal data mining, and social sensing.

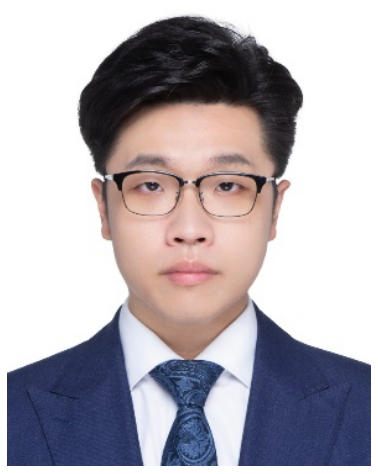

**Zhenghong Wang** received his Ph.D. in Geographic Information Systems at the School of Earth and Space Sciences, Peking University. He is currently an Associate Professor in the College of Computer and Data Science at Fuzhou University. His research interests include spatio-temporal data mining, social sensing, and artificial intelligence.

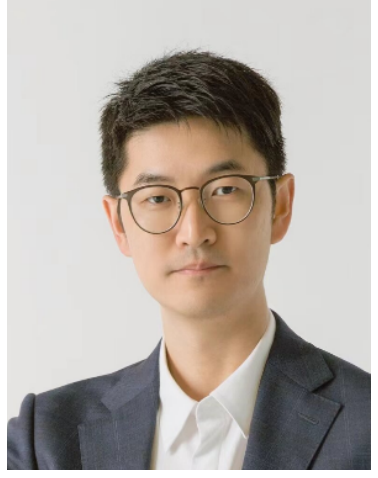

**Fan Zhang** received his B.S. degree of Electronic Engineering from the Beijing Normal University, Zhuhai in 2012 and Ph.D. degree of Earth GeoInformation Science from the Chinese University of Hong Kong in 2017. He is currently an Assistant Professor of GIScience with the Institute of Remote Sensing and Geographical Information Systems, School of Earth and Space Sciences, Peking University. His research interests include Data-driven Urban Studies and Geographical Artificial Intelligence.

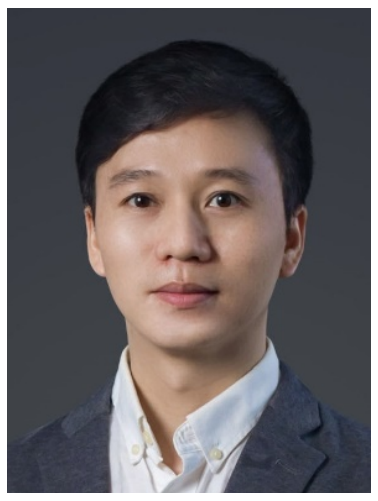

**Chaogui Kang** is a professor of Geographic Information Science at China University University of Geosciences (Wuhan). He was trained in Geography, Data Science, Complex Network, and is leading a dynamic research group – Urban CoLab – with the mission to develop data-driven, human-centric, and cross-cutting methodologies for tackling urban problems from a geospatial perspective. His primary research interest lies in the intersections of Travel Behavior, Built Environment and Social Inequality with the assistance of pervasive urban sensing techniques.

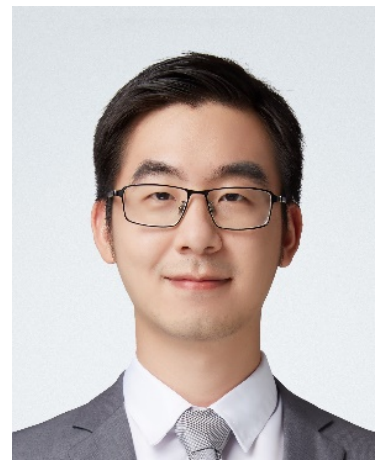

**Sijie Ruan** (Member, IEEE) received the B.E. and Ph.D. degrees from Xidian University, Xi'an, China, in 2017 and 2022, respectively. He is currently an Assistant Professor with the School of Computer Science and Technology, Beijing Institute of Technology (BIT), Beijing, China. Before joining BIT, he was a Visiting Ph.D. Student with Nanyang Technological University, Singapore, and a Research Interns with Microsoft Research Asia, Beijing, and JD Technology, Beijing. His research interests include spatio-temporal data mining and management, volunteered geographic information, and urban computing.

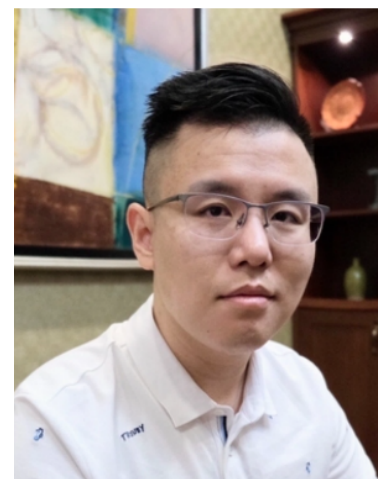

**Di Zhu** is an Assistant Professor of Geographic Information Science in the Department of Geography, Environment, and Society at the University of Minnesota, Twin Cities. His research interests are Geospatial Artificial Intelligence, Spatial Statistics, Social Sensing, and Urban Complexities.

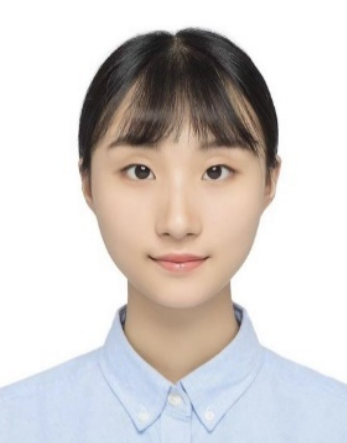

**Chengling Tang** received the B.S. degree in geographic information science from the School of Earth Sciences, Zhejiang University, Hangzhou, Zhejiang, China, in 2022. She is currently pursuing the master's degree with the Institute of Remote Sensing and Geographic Information Systems. Her main research interests include social sensing, urban data mining and GeoAI.

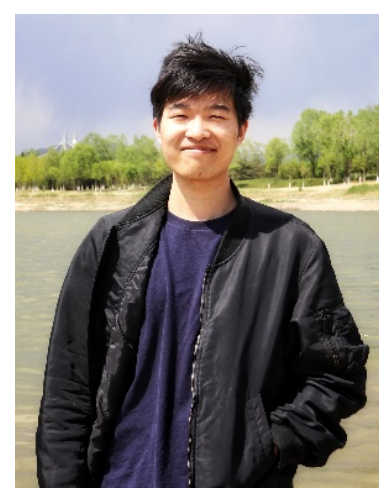

**Zhongfu Ma** is a Ph.D. student in the Department of Geography, Environment, and Society at the University of Minnesota, Twin Cities. His research interest mainly lies in spatial network analysis and spatial process modeling.

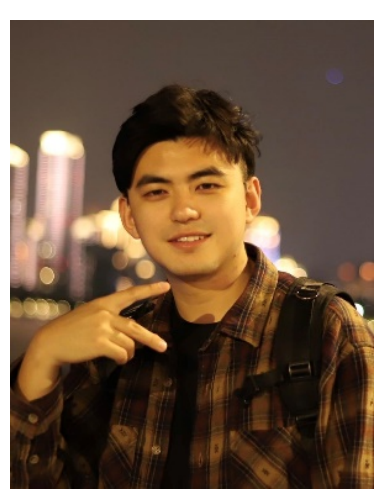

**Weiyu Zhang** received his B.S. degree in Geographic Information Science from Peking University, where he is currently pursuing a Ph.D. His research interests include Geographical Information Systems, Deep Learning (GeoAI), Urban Science, and Urban Data Mining.

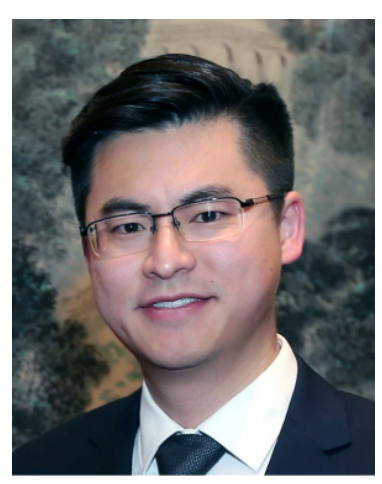

**Yu Zheng** (Fellow, IEEE) is a Vice President of JD.com and a Chief Data Scientist with JD Technology, Beijing, China. He also leads the Intelligent Cities Business Unit as the President and serves as the Managing Director of JD Intelligent Cities Research, Beijing. Before joining JD Technology, he was a Senior Research Manager with Microsoft Research, Beijing. He is also a Chair Professor with Shanghai Jiao Tong University, Shanghai, China, and an Adjunct Professor with Hong Kong University of Science and Technology, Hong Kong. His research interests include big data analytics, spatio-temporal data mining, machine learning, and artificial intelligence. Dr. Zheng is an ACM Distinguished Scientist.

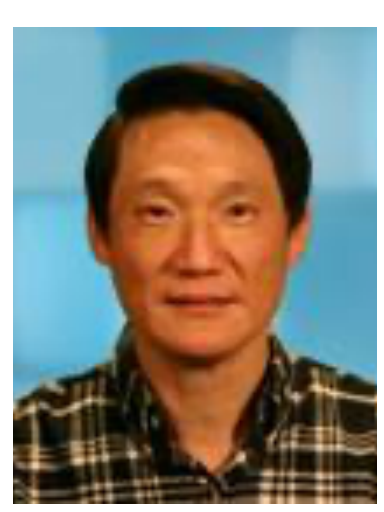

**Philip S. Yu** (Life Fellow, IEEE) is a Distinguished Professor in Computer Science at the University of Illinois at Chicago and holds the Wexler Chair in Information Technology. Before joining UIC, Dr. Yu was with IBM, where he was manager of the Software Tools and Techniques department at the Watson Research Center. His research interest is on big data, and artificial intelligence, including data mining, database and privacy. He has published more than 1,600 papers in refereed journals and conferences. He holds or has applied for more than 300 US patents. Dr. Yu is a Fellow of the ACM and the IEEE. Dr. Yu is the recipient of ACM SIGKDD 2016 Innovation Award for his influential research and scientific contributions on *mining, fusion and anonymization of big data*, the IEEE Computer Society's 2013 Technical Achievement Award for "*pioneering and fundamentally innovative contributions to the scalable indexing, querying, searching, mining and anonymization of big data*", and the Research Contributions Award from IEEE Intl. Conference on Data Mining (ICDM) in 2003 for his pioneering contributions to the field of data mining. He also received the VLDB 2022 Test of Time Award, ACM SIGSPATIAL 2021 10-year Impact Award, WSDM 2020 Honorable Mentions of the Test of Time Award, ICDM 2013 10-year Highest-Impact Paper Award, and the EDBT Test of Time Award (2014). He was the Editor-in-Chiefs of ACM Transactions on Knowledge Discovery from Data (2011-2017) and IEEE Transactions on Knowledge and Data Engineering (2001-2004).

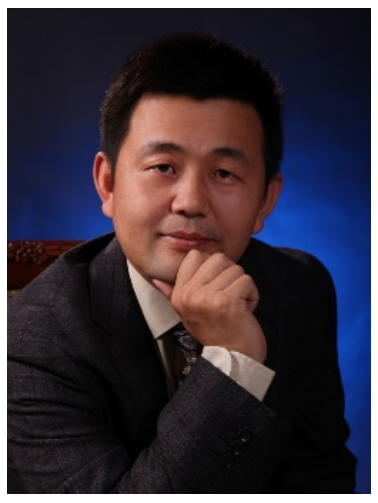

**Yu Liu** received the B.S. and M.S. degrees from the School of City and Environment in 1994 and 1997, respectively, and the Ph.D. degree from the School of Computer Science and Technology, Peking University, Beijing, China, in 2003. He is currently a Boya Professor of GIScience with the Institute of Remote Sensing and Geographical Information Systems, School of Earth and Space Sciences, Peking University. His research interests include humanities and social science based on big geodata.